\documentclass[10pt,twocolumn,letterpaper]{article}

\usepackage{iccv}              

\usepackage{thmtools}
\usepackage{multirow}
\usepackage{subcaption}
\usepackage{algorithm}
\usepackage{algpseudocode}
\usepackage{booktabs}
\usepackage{bbm}
\usepackage{amsmath}
\usepackage{amssymb}
\usepackage{array}
\usepackage{float}
\usepackage[colorinlistoftodos,prependcaption,textsize=tiny]{todonotes}
\usepackage{nicefrac}

\usepackage{multirow}

\definecolor{iccvblue}{rgb}{0.21,0.49,0.74}
\usepackage[pagebackref,breaklinks,colorlinks,allcolors=iccvblue]{hyperref}

\def\paperID{8158} 
\def\confName{ICCV}
\def\confYear{2025}

\title{Guiding Noisy Label Conditional Diffusion Models with \\
Score-based Discriminator Correction}

\author{
\hspace{-3mm}Dat Nguyen Cong\\
\hspace{-3mm}FPT Software AI Center\\
{\hspace{-3mm}\tt\small datnc13@fpt.com}
\and
\hspace{-3mm}Hieu Tran Bao \\
\hspace{-3mm}FPT IS AI R\&D Center \\
{\hspace{-3mm}\tt\small hieutb2@fpt.com
}
\and\hspace{-4mm}
Tung Hoang-Thanh\\
\hspace{-4mm}VNU University of Engineering and Technology\\
{\hspace{-4mm}\tt\small htt210@gmail.com}
}
\begin{document}
\maketitle
\begin{abstract}
Diffusion models have gained prominence as state-of-the-art techniques for synthesizing images and videos, particularly due to their ability to scale effectively with large datasets. 
Recent studies have uncovered that these extensive datasets often contain mistakes from manual labeling processes. 
However, the extent to which such errors compromise the generative capabilities and controllability of diffusion models is not well studied. 
This paper introduces Score-based Discriminator Correction (SBDC), a guidance technique for aligning noisy pre-trained conditional diffusion models. 
The guidance is built on discriminator training using adversarial loss, drawing on prior noise detection techniques to assess the authenticity of each sample.
We further show that limiting the usage of our guidance to the early phase of the generation process leads to better performance.
Our method is computationally efficient, only marginally increases inference time, and does not require retraining diffusion models.
Experiments on different noise settings demonstrate the superiority of our method over previous state-of-the-art methods.

\end{abstract}    
\section{Introduction}
\label{sec:intro}

\begin{figure}[t!]
    \centering
    \includegraphics[width=0.45\textwidth]{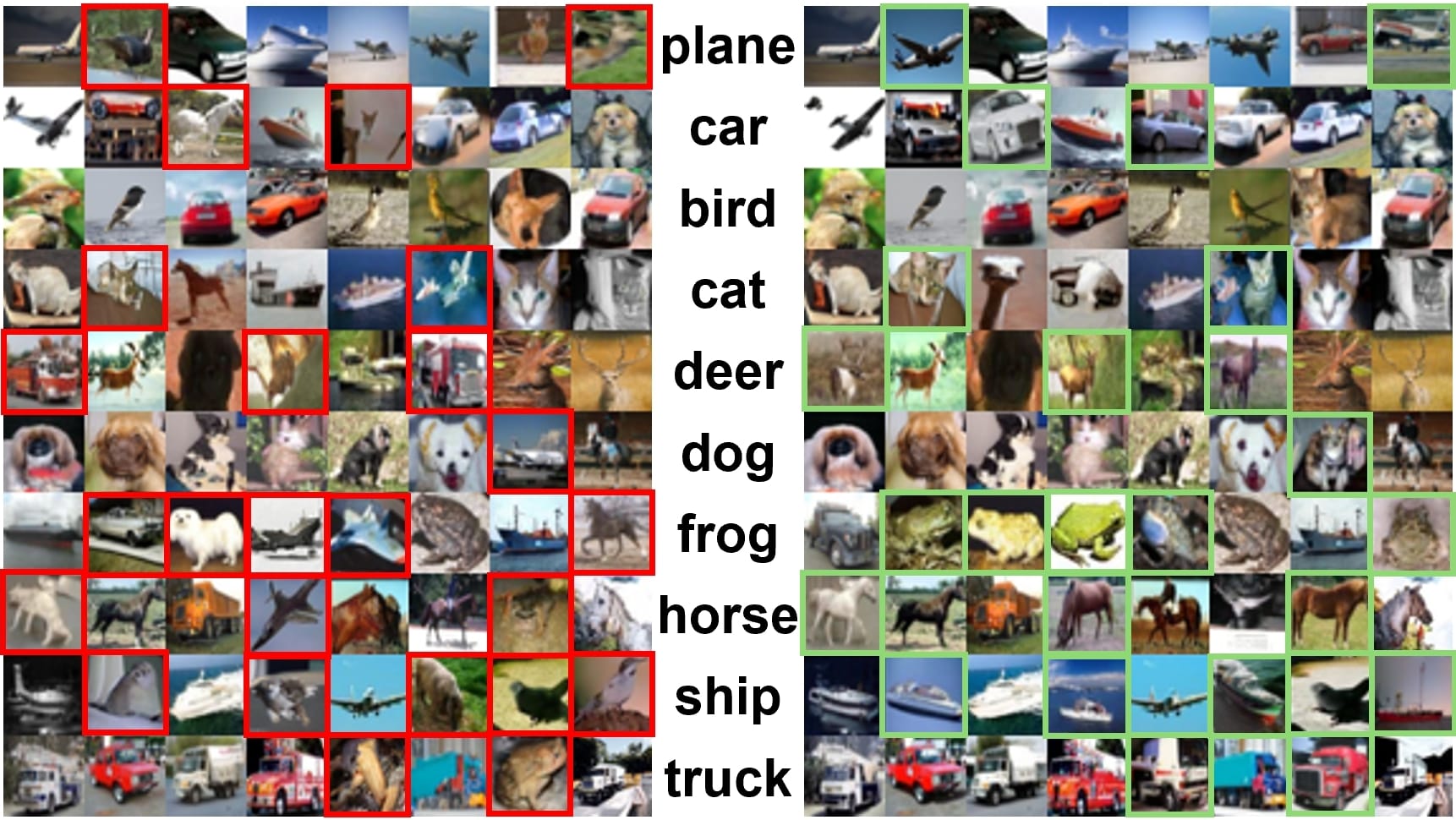}
    \vspace{-2mm}
    \caption{The uncurated generated images of EDM (left) and SBDC (right) on the CIFAR-10 dataset with 50\% symmetric noise. We highlight erroneous images their and corrections in red and green squares, respectively.}
    \vspace{-6mm}
    \label{fig:edm-fig}
\end{figure}

\hspace{0.9em} Diffusion models \cite{pmlr-v37-sohl-dickstein15, ho2020denoising} have emerged as state-of-the-art techniques for generating highly realistic and controllable images and videos. 
Their success is largely attributed to their ability to scale effectively with massive datasets \cite{schuhmann2022laion, wang2022diffusiondb, bain2021frozen}. 
However, these large and often uncurated datasets come with significant challenges, including mislabeled, ambiguous \cite{beyer2020we}, and unsafe content \cite{thiel2023identifying}. 
As a result, models trained on such data may produce low-quality, irrelevant, or even nonsensical outputs.  
These data quality issues are prevalent across both labeled datasets like ImageNet \cite{deng2009imagenet} and multi-modal datasets such as LAION-5B \cite{schuhmann2022laion}, negatively impacting the performance of generative models. 
Manually addressing these issues
is an extremely costly and nearly infeasible task.

Noisy datasets can cause conditional generative models to generate lower-quality images that do not align with the conditions (Fig.~\ref{fig:edm-fig}).
This paper focuses on methods for correcting the outputs of generative models trained on noisy datasets. 
Existing approaches often require retraining generative models with modified loss functions (and/or) on modified datasets.
Works on noisy conditional generation rely on transition matrix estimation techniques \cite{kye2022learning,yao2020dual,zhang2021learning,bae2022noisy}, which adjust the objective function to mitigate the effects of noisy labels \cite{kaneko2019label,na2024labelnoise}.
This approach involves multi-stage training processes, where errors and biases from one stage can propagate to the next.
Alternatively, noise detection methods \cite{cheng2021learninginstancedependentlabelnoise,zhu2022detecting} identify and remove noisy data from the training set and then retrain generative models on the cleaned datasets.
Retraining large-scale generative models is extremely expensive and time-consuming, rendering the above methods impractical for large-scale models.
\begin{figure*}[t!]
    \centering
    \includegraphics[width=\linewidth]{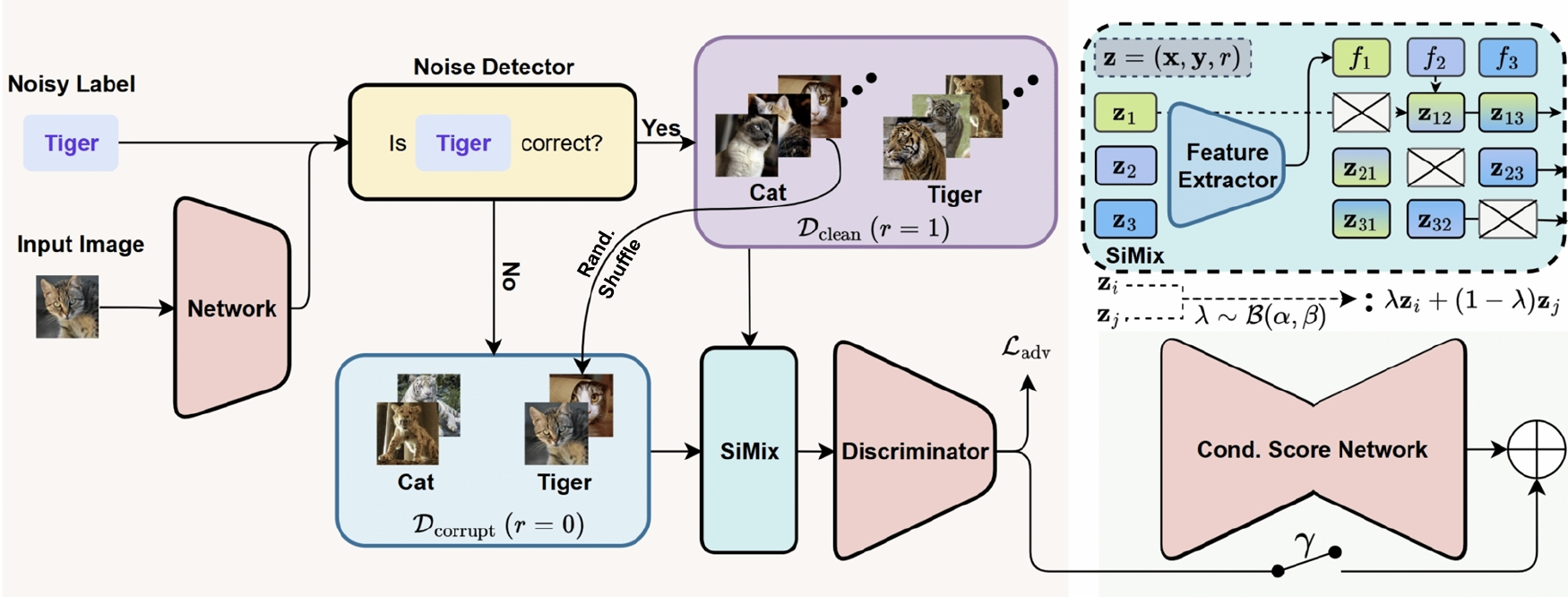}
    \caption{The overall training (left) and inference (bottom right) procedure of the proposed approach. The solid curve from $\mathcal{D}_{\text{clean}}$ to $\mathcal{D}_{\text{corrupt}}$ indicates we randomly shuffle the pseudo-clean label to the corrupted label. 
    In \textbf{SiMix}, the nearest neighbor in the feature space is selected to mix with the current input. 
    The discriminator is narrowly applied through a $\gamma$-gate during the inference phase.}
    \label{fig:pipeline}
    \vspace{-3mm}
\end{figure*}

To address this challenge, we propose Score-based Discriminator Correction (SBDC), an inference-time method that enhances the robustness of pre-trained conditional diffusion models against label noise.
We train a \emph{small} discriminator and use it to guide the generation of the pre-trained conditional diffusion model (\cref{fig:edm-fig}).
Our method does not require retraining pre-trained generative models and only incurs a small computational overhead at inference time.
We show theoretically and empirically that correction guidance can be derived from a discriminator distinguishing correct from incorrect labels.

Our approach is related to Noise-contrastive estimation (NCE) \cite{gutmann2010noise}, which differentiates real data from fake data for unnormalized models. 
Taking advantage from it, we utilize prior works on noise detection and streamline the training process.
The noise detection process and discriminator training are fast and, thus, are highly suitable for inference time methods. 
The generation process can be further optimized for inference time efficiency using soft guidance (\cref{sec:method}). 
Through extensive experiments on multiple datasets—with varying annotation sizes, complexity, and noise levels, we demonstrate that our method consistently outperforms the state-of-the-art Transition-aware Weighted Denoising Score Matching (TDSM) \cite{na2024labelnoise} across several metrics, particularly on real-world and diverse-class datasets. 
We found that TDSM struggles with high-intensity noise, often generating images with incorrect labels. 
In contrast, our approach effectively corrects these errors, significantly enhancing overall performance (Fig.~\ref{fig:tdsm-fig}).

\noindent{\textbf{Our contributions}} 
\begin{itemize}
    \item We propose Score-based Discriminator Correction to align pre-trained noisy conditional diffusion models at inference time. 
    \item We empirically demonstrate that class instability usually 
    happens during the early and middle phase of the sampling process, allowing us to apply our method to a fraction of the sampling process.
    \item Experiments show that SBDC outperforms TDSM, particularly in enhancing intra-class quality, handling high-intensity noise, and is efficient in both post-process training and generation phases.
\end{itemize}

\section{Preliminary}
\label{sec:noisy_cond}
\hspace{0.9em} This section provides a concise overview of diffusion models and the problem of learning with noisy labels.

\noindent\textbf{Definition.} We denote the train set as $\mathcal{D}=\{(\textbf{x}_i,\textbf{y}_i)\}_{i=1}^{n}$, where $\textbf{x}_i\in \mathcal{X}$ is the data instance and $\textbf{y}_i\in\mathcal{Y}$ is the true label of the $i^{th}$ sample. 
$\tilde{\textbf{y}}_i \in \mathcal{Y}$ denotes the observed (and potentially incorrect) label of the $i^{th}$ sample. 
A conditional generative dataset consists of i.i.d. samples from the joint probability distribution $P$ over $\mathcal{X}\times\mathcal{Y}$.
During training, only $\tilde{\mathcal{D}}=\{(\textbf{x}_i,\tilde{\textbf{y}}_i)\}_{i=1}^{n}$ is observed and $\tilde{\mathcal{D}}$ is potentially different from $\mathcal{D}$.
We also denote $\mathcal{D}_{r} = \{(\mathbf{x}, \mathbf{y}_r)\in \tilde{\mathcal{D}}|\mathbf{y}_r = \mathbf{y}\}$ and $p_r\sim\mathcal{D}_{r}$ as the clean distribution, $\mathcal{D}_{f} = \{(\mathbf{x}, \mathbf{y}_f)\in \tilde{\mathcal{D}}|\mathbf{y}_f \neq \mathbf{y}\}$ and $p_f\sim\mathcal{D}_{f}$ as the corrupt distribution.

\noindent \textbf{Diffusion Models.} 
\citet{anderson1982reverse,song2020score} represent the diffusion process through first-order stochastic differential equations. 
The forward diffusion process in \cref{eq:forward_sde} models the time-dependent displacement $\{\mathbf{x}_t\}_{t=0}^T$ 
\begin{equation}
    d\mathbf{x}_t = \text{f}(\mathbf{x}_t, t)dt + \text{g}(t)d\text{w}_t,
    \label{eq:forward_sde}
\end{equation}
where $\text{f}(\cdot,\cdot):\mathbb{R}^d\times\mathbb{R}\to\mathbb{R}^d$ is the drift, $\text{g}(\cdot):\mathbb{R}\to\mathbb{R}$ is the diffusion coefficient, and $\text{w}_t$ is a $d$-dimensional Brownian motion.   
This process gradually transforms the data distribution $q(\mathbf{x}_0)$ to a simple distribution $q(\mathbf{x}_T)$, typically a Gaussian distribution.
The forward process has a unique reverse-time diffusion process which converts $\mathbf{x}_T$ to $\mathbf{x}_0$
\begin{align}
    d\mathbf{x}_t = \left[\text{f}(\mathbf{x}_t, t) - \text{g}^2(t)\nabla_{\mathbf{x}_t}\log p(\mathbf{x}_t)\right] dt + \text{g}(t) d\bar{\text{w}}_t, \label{eq:reverse_sde}
\end{align}
where $\bar{\text{w}}_t$ is a $d$-dimensional reverse-time Brownian motion, $\nabla_{\mathbf{x}_t}\log p(\mathbf{x}_t)$ represents the score of the data instance.
Learning the score function $s_\theta(\mathbf{x}_t,t) \approx \nabla_{\mathbf{x}_t}\log p(\mathbf{x}_t)$ allows sampling from the data distribution through the reverse process.
Since it is intractable,~\citet{song2019generative} trains the score network with the per-sample data score $\nabla_{\mathbf{x}_t}\log p(\mathbf{x}_t|\mathbf{x}_0)$ as the new target, resulting in the denoising score matching (DSM) loss
\begin{align}
    &L_{\text{DSM}}(\theta,t) = \nonumber\\&\mathbb{E}_{p(\mathbf{x}_0)}\mathbb{E}_{p(\mathbf{x}_t|\mathbf{x}_0)} \Big[\lvert|\nabla_{\mathbf{x}_t}\log p(\mathbf{x}_t|\mathbf{x}_0) - s_{\theta}(\mathbf{x}_t,t) \rvert|_2^2\Big].
\end{align}

\noindent\textbf{Conditional Diffusion Models with Label Noise.}
In conditional generation, the target distribution is $p(X|Y)$, where $X$ is generated from the label $Y\in \{1,...,K\}$.
During training, $Y$ is added to the denoising network as auxiliary information to guide the generation process towards the distribution of the target sample. 
However, in real-world settings, only the noisy label $\tilde{Y}$ is observed. 
Training the score network with the objective
\begin{align}
     &L_{\text{DSM}}(\theta,t) = \\&\mathbb{E}_{p(\mathbf{x}_0,\tilde{\mathbf{y}})}\mathbb{E}_{p(\mathbf{x}_t|\mathbf{x}_0,\tilde{\mathbf{y}})} \Big[\lvert| \nabla_{\mathbf{x}_t}\log p(\mathbf{x}_t|\mathbf{x}_0,\tilde{\mathbf{y}})- s_{\theta}(\mathbf{x}_t,\tilde{\mathbf{y}},t) \rvert|_2^2\Big],\nonumber    \label{eq:noisy_obj}
\end{align}
eventually makes $s_{\theta}(\mathbf{x}_t,\tilde{\mathbf{y}},t)$, or $s_{\theta}(\mathbf{x}_t,\tilde{\mathbf{y}})$ in short, learn the noisy distribution $p(X|\tilde{Y})$.

\noindent \textbf{Refining Generative Process in Inference Time.}
As recent models continue to grow in size, retraining to reduce approximation errors becomes impractical.
Discriminator Guidance \cite{kim2022refining} refines the score function using discriminator signals to distinguish real from synthesized samples
\begin{align}
    \nabla_{\mathbf{x}_t}\log p_\theta^\phi(\mathbf{x}_t)&=
    s_\theta(\mathbf{x}_t) + w\log\frac{D_\phi^t(\mathbf{x}_t)}{1-D_\phi^t(\mathbf{x}_t)},
\end{align}
where $w$ is a constant guidance weight.
Sampling with the new score reduces the model's approximation error.
We also use a discriminator, but train the discriminator solely on real data for direct alignment estimation \(\nabla\log\frac{p(\mathbf{x}_t|\mathbf{y})}{p(\mathbf{x}_t|\mathbf{\tilde{y}})}\).

\section{Methodologies}
\label{sec:method}
\hspace{0.9em} In this section, we present our approach to training the discriminator and correcting the score estimation during inference with it. 
In \cref{subsec:motivation}, we study the effect of training bias on the deviation of generation trajectory from the correct path. 
\cref{subsec:dg} introduces the discriminator correction to guide the target distribution trajectory.
To boost the inference time, we limit the discriminator usage at a specific interval (\cref{subsec:limited-int}).
The practical training implementation is presented in \cref{subsec:mixup}.
We illustrate the training and inference procedure in \cref{fig:pipeline}.

\subsection{Noisy Conditional Generative Behavior}
\label{subsec:motivation}
\begin{figure}[t!]
    \centering
    \begin{subfigure}[h]{0.575\linewidth}
        \centering
        \includegraphics[width=\textwidth]{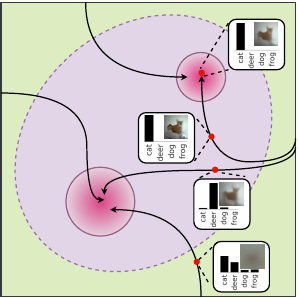} 
        \caption{Conditional target field}\label{subfig:denoise_path}
    \end{subfigure}
    \hfill
    \begin{subfigure}[t]{0.415\linewidth}
    \begin{subfigure}{\linewidth} 
        \centering
        \includegraphics[width=\textwidth]{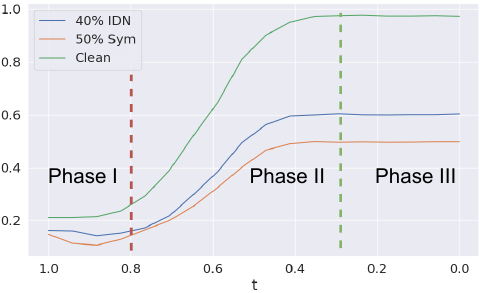} 
        \caption{Confidence}
        \label{subfig:confidence}
    \end{subfigure}
    \vspace{1em}
    \begin{subfigure}{\linewidth}
        \centering
        \includegraphics[width=\textwidth]{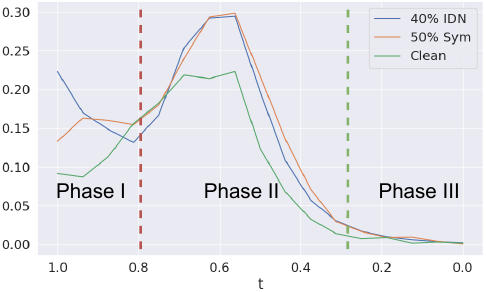} 
        \caption{Instability}
        \label{subfig:instability}
    \end{subfigure}
    \end{subfigure}
    \caption{(a): Illustration of the three phases in a two-mode distribution.
    (b,c): Estimated $C(t)$ and $I(t)$. 
    We use 18-step Heun sampling (normalizing the maximum value to 1) and the reverse process of 2k samples for each noise setting. }
    \label{fig:analysis}
\end{figure}
\hspace{0.9em} During the generation process, the diffusion model first determines the update direction by estimating the fully denoised sample $\mathbf{x}_\theta(\mathbf{x}_t,\mathbf{y})$ from $\mathbf{x}_t$, and then updates the next state following \cref{eq:reverse_sde}.
The updated direction can be viewed as a Monte-Carlo estimator $\nabla_{\mathbf{x}_t} \log p_\theta(\mathbf{x}_t|\mathbf{y}) = \mathbb{E}_{p_\theta(\mathbf{x}_0 | \mathbf{x}_t,\mathbf{y})} \left[ \nabla_{\mathbf{x}_t} \log p(\mathbf{x}_t | \mathbf{x}_0) \right]\approx \frac{1}{n} \sum_{i=1}^n \nabla_{\mathbf{x}_t} \log p(\mathbf{x}_t| \mathbf{x}_\theta^{(i)})$.
Typically, $\mathbf{x}_\theta^{(i)}$ is sampled from $p_\theta(\cdot|\mathbf{x}_t,\mathbf{y})$ and $n=1$, meaning the \textit{influence} of a single pair $(\mathbf{x}_0, \mathbf{y})$ on $\mathbf{x}_t$.
This results in high bias to the individual target sample during sampling, which is crucially harmful in the mislabeled cases. 

We quantitatively characterize the 
variations of individual targets $\mathbf{x}_\theta(\mathbf{x}_t,\mathbf{y})$ at time $t$ with the \textit{confidence} ($C$) and the \textit{instability} ($I$) generation probabilities, defined as
\begin{align}
    C(t) &= \mathbb{P}\Big[ f(\mathbf{x}_\theta(\mathbf{x}_t, \mathbf{y})) = \mathbf{y}\Big], \label{eq:conf_f} \\
    I(t) &= \mathbb{P}\Big[ f(\mathbf{x}_{\theta}(\mathbf{x}_t, \mathbf{y})) \neq f(\mathbf{x}_{\theta}(\mathbf{x}_{t-1}, \mathbf{y}))\Big], \label{eq:insta_f}
\end{align}
where $f(\cdot)$ predicts the class of the denoised sample $\mathbf{x}_\theta$ at step $t$. 
In practice, $f(\cdot)$ is often a large pre-trained classifier, since it remains robust to blurry intermediate samples $\mathbf{x}_\theta$ with high confidence (see \cref{subfig:denoise_path}).

As shown in \cref{subfig:denoise_path}, the sampling process is partitioned into marginalized, conditional, and fine-grained phases (Phase I$\sim$III respectively). 
Intuitively, $I(t)$ peaks in the conditional phase (Phase II), where multiple modes in the noisy data distribution have comparable influences on the perturbation identity, resulting in unstable targets.
In Phase I, $t$ is large and $C(t)$ is low because the labels are mostly discarded from the conditional distribution: $\lim_{t\to \infty}p(\mathbf{x}_t|\mathbf{x}_0,\mathbf{y})=\lim_{t\to \infty}p(\mathbf{x}_t|\mathbf{x}_0) = p(\mathbf{x}_\infty) \approx p(\mathbf{x}_T)$. 
In Phase III, the posterior $p(\mathbf{x}_0|\mathbf{x}_t, \mathbf{y})$ concentrates around a single target, thus $C(t)$ solely reflects the noise rate with low $I(t)$. \cref{subfig:confidence,subfig:instability} empirically validate this argument.
They show the estimated $C(t)$ and $I(t)$ for samples generated from models trained on perturbed CIFAR-10 dataset \cite{krizhevsky2009learning}.
Essentially, $I(t)$ peaks in Phase II, demonstrating a large class variation. 
Since the variation drops as the sampling proceeds, this error remains till the end.
Our method for measuring $C(t)$ and $I(t)$ is described in \cref{alg:conf_inst} (see \cref{sec:conf_inst}).

In \cref{subsec:dg}, we intervene in the sampling trajectory by using additional auxiliary signals to keep the trajectory on the correct path.
From the analysis in \cref{fig:analysis}, we limit the guidance signal to the conditional phase only and present a simple strategy to do so in \cref{subsec:limited-int}.

\subsection{Score-Based Discriminator Correction}
\label{subsec:dg}
\hspace{0.9em} Assuming the score network $s_{\theta}(\mathbf{x}_t,\tilde{\mathbf{y}})$ perfectly learns the noisy distribution $p(\mathbf{x}_t, \tilde{\mathbf{y}})$, we recover the clean distribution generative process by a simple modification
\begin{align}
    \nabla_{\mathbf{x}_t}\log p(\mathbf{x}_t| \mathbf{y})=\nabla_{\mathbf{x}_t}\log p_\theta(\mathbf{x}_t| \tilde{\mathbf{y}})+\nabla_{\mathbf{x}_t}\log\frac{ p(\mathbf{x}_t| \mathbf{y})}{p_{\theta}(\mathbf{x}_t| \tilde{\mathbf{y}})}
\end{align}
where $p_{\theta}$ is the distribution learned from $s_{\theta}$.
Following \citet{kim2022refining}, we approximate the intractable term $\nabla_{\mathbf{x}_t}\log\nicefrac{ p(\mathbf{x}_t| \mathbf{y})}{p_{\theta}(\mathbf{x}_t| \tilde{\mathbf{y}})}$ with the gradient from a time-dependent discriminator $D_\theta^t(\mathbf{x}_t,\mathbf{y})$, which is trained to discriminate between real and synthetic data.

In our case, we define real as clean data sampled from $p_r$ and fake as corrupted data sampled from $p_f$.
From the definition in \cref{sec:noisy_cond}, $p_r$ and $p_f$ can be seen as the truncated  distribution of $p(\mathbf{x},\mathbf{y}|\tilde{\mathcal{D}})$.
Hence, we formulate the discriminator's training objective as the time-weighted binary cross-entropy loss given clean and corrupt samples:
\begin{align}
\mathcal{L}_{\text{adv}}&= \mathbb{E}_{t,(\mathbf{x},\mathbf{y})\sim p_r,\mathbf{x}_t}\left[-\log D_\theta^t(\mathbf{x}_t,\mathbf{y})\right] \nonumber\\&+ \mathbb{E}_{t,(\mathbf{x},\mathbf{y})\sim p_{f},\mathbf{x}_t}\left[-\log(1-D_\theta^t(\mathbf{x}_t,\mathbf{y}))\right].
\label{eq:dg_obj}
\end{align} 
Following this objective, we use a noise detector to filter clean and corrupt data from the training set and assign each instance in each set the ``verified label'' $r$, where $r=1$ corresponds to the pseudo-clean instance and vice versa.
\cref{thrm:dg2ratio} shows that when the noise rate is not overwhelming, training the discriminator with the objective in \cref{eq:dg_obj} allows the estimation of the log-likelihood ratio.
\begin{restatable}{theorem}{theorem1}
    Let $(\cdot,\mathbf{y}_r)\sim p_r,\ (\cdot,\mathbf{y}_f)\sim p_f$, and $D_\theta(\cdot,\cdot)=\sigma(g_\theta(\cdot,\cdot))$ be the logistic function of the discriminator. 
    Assume $g$ satisfies the Lipschitz condition: there exists $L>0$ such that for all $t\in[\epsilon,T]$, $\mathbf{x}$, $\mathbf{z}$, and $\mathbf{y}$, we have $||g(\mathbf{x},\mathbf{y},t)-g(\mathbf{z},\mathbf{y},t)||_2\leq L||\mathbf{x}-\mathbf{z}||_2$. 
    Then, given an optimally trained $D_{\theta^*}$, we have 
    \begin{align}
\underset{{\mathbf{x}_t,\mathbf{y},\tilde{\mathbf{y}},\atop \mathbf{y}_r,\mathbf{y}_f}}{\mathbb{E}}&\Big[\Big\|\nabla_{\mathbf{x}_t}\log\frac{D_{\theta^*}(\mathbf{x}_t,\mathbf{y}_r)}{D_{\theta^*}(\mathbf{x}_t,\mathbf{y}_f)}-\nabla_{\mathbf{x}_t}\log\frac{p(\mathbf{x}_t|\mathbf{y})}{p(\mathbf{x}_t|\tilde{\mathbf{y}})}\Big\|_2^2 \Big] \nonumber\\ &\leq L+\underset{{\mathbf{x}_t,\mathbf{y},\tilde{\mathbf{y}}}}{\mathbb{E}} \Big[\big\| \nabla_{\mathbf{x}_t}\log\frac{p(\mathbf{x}_t|\mathbf{y})}{p(\mathbf{x}_t|\tilde{\mathbf{y}})}\big\|_2^2\Big]
    \label{eq:dg_ratio}
\end{align}
    \label{thrm:dg2ratio}
\end{restatable}

\hspace{0.9em} From \Cref{thrm:dg2ratio}, we ideally assign an auxiliary label $r$ for each instance utilizing previous noise detection methods in classification tasks \cite{kim2024learning,northcutt2021confident,pleiss2020identifying,zhu2022detecting}.  
The proof is presented in \cref{sec:proof}.  

\begin{algorithm}[t!]
\caption{Discriminator Training}
\begin{algorithmic}[1]
\State Construct $\mathcal{D}_r = \{\mathbf{z}_i=(\mathbf{x}_i, \mathbf{y}_i, 1)\}_{i=1}^M$ and $\mathcal{D}_f = \{\mathbf{z}_j=(\mathbf{x}_j, \mathbf{y}_j, 0)\}_{j=1}^N$ from noise detection method.

\While{not converged}
    \State Sample $\mathcal{Z}=\{\mathbf{z}_i^c\}_{i=1}^{m}\cup\{\mathbf{z}_i^l\}_{i=1}^{n}$ from $\mathcal{D}_r\cup\mathcal{D}_f$
    \State Random shuffle $\{\mathbf{z}_i^c\}_{i=1}^{k}\leftarrow\{(\mathbf{x}_i, \tilde{\mathbf{y}}_i, 0)\}_{i=1}^{k}$ 
    \Comment{Pseudo-clean Shuffle}
    \State $\mathcal{Z}\leftarrow\mathbf{Simix(\mathcal{Z})}$
    \Comment{SiMix Augmentation}
    \State Sample $t_1, \ldots, t_{m+n}$ from $[1..T]$
    \For{$\epsilon_i \sim \mathcal{N}(0, \mathbf{I}), \forall i = 1, \ldots, m+n$}
    \State $\mathbf{x}_i^{t_i} \leftarrow \sqrt{\bar{\alpha}_{t_i}} \mathbf{x}_i + \sqrt{1 - \bar{\alpha}_{t_i}} \epsilon_i$ 
    \EndFor
    
    \State Calculate $\hat{\mathcal{L}}_\phi \leftarrow - \sum_{i=1}^{m+n} r_i \log D_\phi^{t_i}(\mathbf{x}_i^{t_i}, \mathbf{y}_i) + (1 - r_i) \log (1-D_\phi^{t_i}(\mathbf{x}_i^{t_i}, \mathbf{y}_i)) $
    \State Update $\phi \leftarrow \phi - \frac{\partial \hat{\mathcal{L}}_\phi}{\partial \phi}$
\EndWhile
\end{algorithmic}
\label{alg:dg-train}
\end{algorithm}
\begin{algorithm}[t!]
\caption{\textbf{SiMix}($\mathcal{Z}$)}
\label{alg:simix}
\begin{algorithmic}[1]
\State $\mathcal{F} = \{f_i \mid f_i = \text{Encode}(\mathbf{x}_i), \forall \mathbf{z}_i:=(\mathbf{x}_i, \mathbf{y}_i, r_i) \in \mathcal{Z}\}$
\State Sample $\lambda_1, \ldots, \lambda_B$ from $\text{Beta}(\alpha, \alpha)$
\For{$i$ in $[1..B]$} 
    \State $\mathbf{z}_i \leftarrow \lambda_i \mathbf{z}_i + (1-\lambda_i) \mathbf{z}_{\underset{{j\in[1..B]}}{\arg\min} \|f_i - f_j\|_2}$
\EndFor
\State \textbf{Return} new $\mathcal{Z}$
\end{algorithmic}
\end{algorithm}

\begin{table*}[t!]
\setlength{\tabcolsep}{1.pt} 
\centering
\small
\begin{tabular}{lp{3mm}@{\hspace{1mm}} >{\centering\arraybackslash}m{1.61cm} >{\centering\arraybackslash}m{1.7cm} >{\centering\arraybackslash}m{1.7cm}  p{0.005mm} >{\centering\arraybackslash}m{1.61cm} >{\centering\arraybackslash}m{1.7cm} >{\centering\arraybackslash}m{1.7cm}
p{0.005mm} >{\centering\arraybackslash}m{1.61cm} >{\centering\arraybackslash}m{1.7cm} >{\centering\arraybackslash}m{1.65cm}
}
\hline
{CIFAR-10} &  & EDM & TDSM & SBDC & & EDM & TDSM & SBDC & & EDM & TDSM & SBDC  
\\
\hline
\multirow{2}{*}{Metrics} &  & \multicolumn{3}{c}{$\text{Aymmetric}^{\ddagger}$} & & \multicolumn{3}{c}{Symmetric} & & \multicolumn{3}{c}{Symmetric} \\ 
\cline{3-5} \cline{7-9} \cline{11-13}
\noalign{\vskip 2pt}
& & \multicolumn{3}{c}{20\%} & & \multicolumn{3}{c}{50\%} & & \multicolumn{3}{c}{80\%}  \\
\hline
\noalign{\vskip 2pt}
FID & ($\downarrow$) & $\mathbf{1.96}$\footnotesize{ $\pm$ 0.02} & $\underline{2.36}$\footnotesize{ $\pm$ 0.02} & ${2.49}$\footnotesize{ $\pm$ 0.01} & & $\mathbf{2.07}$\footnotesize{ $\pm$ 0.02} & $2.43$\footnotesize{ $\pm$ 0.03} & $\underline{2.24}$\footnotesize{ $\pm$ 0.01} & & $\underline{2.15}$\footnotesize{ $\pm$ 0.04} & $\mathbf{2.25}$\footnotesize{ $\pm$ 0.02} & ${2.30}$\footnotesize{ $\pm$ 0.21} \\
IS & ($\uparrow$) & $9.95$\footnotesize{ $\pm$ 0.01} & $\underline{10.04}$\footnotesize{ $\pm$ 0.08} & $\mathbf{10.06}$\footnotesize{ $\pm$ 0.03} & & $9.69$\footnotesize{ $\pm$ 0.06} & $\underline{9.84}$\footnotesize{ $\pm$ 0.03} & $\mathbf{9.87}$\footnotesize{ $\pm$ 0.03} & & $9.67$\footnotesize{ $\pm$ 0.06} & $\underline{9.76}$\footnotesize{ $\pm$ 0.02} & $\mathbf{9.71}$\footnotesize{ $\pm$ 0.11}  \\
Density & ($\uparrow$) & $103.9$\footnotesize{ $\pm$ 0.40} & $\underline{115.9}$\footnotesize{ $\pm$ 0.07} & $\mathbf{118.1}$\footnotesize{ $\pm$ 0.49} & & $103.6$\footnotesize{ $\pm$ 0.42} & $\mathbf{114.0}$\footnotesize{ $\pm$ 1.22} & $\underline{112.8}$\footnotesize{ $\pm$ 0.48} & & $103.2$\footnotesize{ $\pm$ 0.16} & $\mathbf{108.0}$\footnotesize{ $\pm$ 0.47} & $\underline{104.1}$\footnotesize{ $\pm$ 0.46}  \\
Coverage & ($\uparrow$) & $83.7$\footnotesize{ $\pm$ 0.09} & $\mathbf{85.1}$\footnotesize{ $\pm$ 0.11} & $\underline{84.9}$\footnotesize{ $\pm$ 0.04} & & $83.2$\footnotesize{ $\pm$ 0.09} & $\mathbf{84.3}$\footnotesize{ $\pm$ 0.21} & $\underline{84.0}$\footnotesize{ $\pm$ 0.22} & & $82.8$\footnotesize{ $\pm$ 0.13} & $\mathbf{83.9}$\footnotesize{ $\pm$ 0.03} & $\underline{82.7}$\footnotesize{ $\pm$ 0.33}  \\
CW-FID & ($\downarrow$) & $11.3$\footnotesize{ $\pm$ 0.03} & $\underline{10.9}$\footnotesize{ $\pm$ 0.02} & $\mathbf{10.6}$\footnotesize{ $\pm$ 0.02} & & $38.6$\footnotesize{ $\pm$ 0.18} & $\underline{18.2}$\footnotesize{ $\pm$ 0.16} & $\mathbf{15.6}$\footnotesize{ $\pm$ 0.10} & & $71.7$\footnotesize{ $\pm$ 0.09} & $\underline{59.8}$\footnotesize{ $\pm$ 0.15} & $\mathbf{48.2}$\footnotesize{ $\pm$ 0.47}  \\
CW-Den. & ($\uparrow$) & $98.6$\footnotesize{ $\pm$ 0.42} & $\underline{113.1}$\footnotesize{ $\pm$ 0.17} & $\mathbf{114.8}$\footnotesize{ $\pm$ 0.60} & & $66.8$\footnotesize{ $\pm$ 0.87} & $\underline{95.8}$\footnotesize{ $\pm$ 0.97} & $\mathbf{98.1}$\footnotesize{ $\pm$ 0.34} & & $43.0$\footnotesize{ $\pm$ 0.12} & $\underline{52.0}$\footnotesize{ $\pm$ 0.36} & $\mathbf{58.0}$\footnotesize{ $\pm$ 0.11}  \\
CW-Cov. & ($\uparrow$) & $82.3$\footnotesize{ $\pm$ 0.08} & $\mathbf{84.4}$\footnotesize{ $\pm$ 0.07} & $\underline{83.8}$\footnotesize{ $\pm$ 0.02} & & $68.9$\footnotesize{ $\pm$ 0.42} & $\underline{79.5}$\footnotesize{ $\pm$ 0.42} & $\mathbf{79.9}$\footnotesize{ $\pm$ 0.18} & & $44.9$\footnotesize{ $\pm$ 0.71} & $\underline{55.9}$\footnotesize{ $\pm$ 0.62} & $\mathbf{59.1}$\footnotesize{ $\pm$ 0.08} \\
\hline
\rule{0pt}{2pt}
\multirow{2}{*}{Metrics} & & \multicolumn{3}{c}{Instance
} & & \multicolumn{3}{c}{Instance} & & \multicolumn{3}{c}{$\text{Average}^*$} \\
\cline{3-5} \cline{7-9} \cline{11-13}
\noalign{\vskip 2pt}
& & \multicolumn{3}{c}{20\%} & & \multicolumn{3}{c}{40\%} &  & \begin{tabular}{m{1.4cm} <{\centering} m{1.2cm} m{1.3cm} m{1.2cm}}
         EDM & TDSM & SBDC & Clean
      \end{tabular} \\
\hline
\noalign{\vskip 2pt}
FID & ($\downarrow$) & $\mathbf{1.95}$\footnotesize{ $\pm$ 0.02} & $2.53$\footnotesize{ $\pm$ 0.01} & \underline{2.44}\footnotesize{ $\pm$ 0.02} & & $\mathbf{1.97}$\footnotesize{ $\pm$ 0.02} & $\underline{2.20}$\footnotesize{ $\pm$ 0.02} & $2.49$\footnotesize{ $\pm$ 0.01} & & \begin{tabular}{m{1.6cm} <{\centering} m{1.2cm} m{1.2cm} m{1.2cm}}
         \textbf{2.02} & \underline{2.34} & 2.39 & 1.88
      \end{tabular} \\
IS & ($\uparrow$) & $\underline{9.80}$\footnotesize{ $\pm$ 0.05} & $9.73$\footnotesize{ $\pm$ 0.01} & $\mathbf{9.87}$\footnotesize{ $\pm$ 0.03} & & $9.77$\footnotesize{ $\pm$ 0.04} & $\underline{9.85}$\footnotesize{ $\pm$ 0.03} & $\mathbf{9.88}$\footnotesize{ $\pm$ 0.01} & & \begin{tabular}{m{1.6cm} <{\centering} m{1.2cm} m{1.2cm} m{1.2cm}}
         9.78 & \underline{9.84} & \textbf{9.88} & 9.96
      \end{tabular}  \\
Density & ($\uparrow$) & $103.2$\footnotesize{ $\pm$ 0.19} & $\underline{113.1}$\footnotesize{ $\pm$ 0.36} & $\mathbf{119.7}$\footnotesize{ $\pm$ 0.57} & & $102.7$\footnotesize{ $\pm$ 0.53} & $\underline{112.0}$\footnotesize{ $\pm$ 0.18} & $\mathbf{116.6}$\footnotesize{ $\pm$ 0.44} & & \begin{tabular}{m{1.5cm} <{\centering} m{1.2cm} m{1.2cm} m{1.2cm}}
         103.3 & \underline{112.6} & \textbf{114.2} & 104.4
      \end{tabular}  \\
Coverage & ($\uparrow$) & $83.2$\footnotesize{ $\pm$ 0.17} & $\underline{84.0}$\footnotesize{ $\pm$ 0.13} & $\mathbf{84.4}$\footnotesize{ $\pm$ 0.06} & & $83.0$\footnotesize{ $\pm$ 0.19} & $\mathbf{84.4}$\footnotesize{ $\pm$ 0.10} & $\underline{84.1}$\footnotesize{ $\pm$ 0.12} & & \begin{tabular}{m{1.6cm} <{\centering} m{1.2cm} m{1.2cm} m{1.2cm}}
         83.2 & \textbf{84.3} & \underline{84.0} & 83.9
      \end{tabular} \\
CW-FID & ($\downarrow$) & $16.3$\footnotesize{ $\pm$ 0.18} & $\underline{11.9}$\footnotesize{ $\pm$ 0.10} & $\mathbf{11.6}$\footnotesize{ $\pm$ 0.06} & & $29.7$\footnotesize{ $\pm$ 0.05} & $\underline{18.1}$\footnotesize{ $\pm$ 0.04} & $\mathbf{13.6}$\footnotesize{ $\pm$ 0.15} & & \begin{tabular}{m{1.6cm} <{\centering} m{1.2cm} m{1.3cm} m{1.2cm}}
         33.5 & \underline{23.8} & \textbf{19.9} & 9.8
      \end{tabular}  \\
CW-Den. & ($\uparrow$) & $90.1$\footnotesize{ $\pm$ 0.26} & $\underline{106.5}$\footnotesize{ $\pm$ 0.38} & $\mathbf{114.9}$\footnotesize{ $\pm$ 0.19} & & $74.9$\footnotesize{ $\pm$ 0.50} & $\underline{94.2}$\footnotesize{ $\pm$ 0.31} & $\mathbf{106.9}$\footnotesize{ $\pm$ 0.45} & & \begin{tabular}{m{1.6cm} <{\centering} m{1.2cm} m{1.1cm} m{1.2cm}}
         74.7 & \underline{92.3} & \textbf{98.5} & 103.3
      \end{tabular} \\
CW-Cov. & ($\uparrow$) & $79.6$\footnotesize{ $\pm$ 0.13} & $\underline{82.4}$\footnotesize{ $\pm$ 0.23} & $\mathbf{83.1}$\footnotesize{ $\pm$ 0.16} & & $73.5$\footnotesize{ $\pm$ 0.22} & $\underline{80.0}$\footnotesize{ $\pm$ 0.09} & $\mathbf{81.5}$\footnotesize{ $\pm$ 0.20} & & \begin{tabular}{m{1.6cm} <{\centering} m{1.2cm} m{1.2cm} m{1.2cm}}
         69.8 & \underline{76.4} & \textbf{77.5} & 83.5
      \end{tabular} \\
\hline
\end{tabular}
\caption{Performance comparison on CIFAR-10 dataset under various noise conditions. The percentage below noise type represents the noise rate. 
The variance is computed through 3 seeds of measurement.
The best results are in \textbf{bold} and the second best ones are in \underline{underlined}. 
*Average results across five different settings. $\ddagger$ implies SBDC used CL, while others used CORES.}
\label{tab:cifar10_var_results}
\end{table*}
\subsection{Limited Interval Guidance Improves Sample Quality}
\label{subsec:limited-int}
\hspace{0.9em} From the observation in \cref{subsec:motivation}, the instability of the generation process is fuzzy in the marginalized phase (high noise levels), extremely low in the fine-grained phase (low noise levels), and peaks in the conditional phase. 
Hence, we schedule the guidance only in the middle phase. 

\citet{kynkaanniemi2024applying} shows that classifier-free guidance also has a sweet spot for optimal guidance interval. 
Motivated by their schedule, we formulate the guidance as:
\begin{align}
    \nabla_{\mathbf{x}_t}\log p(\mathbf{x}_t|\mathbf{y})&=
    s_\theta(\mathbf{x}_t,\tilde{\mathbf{y}}) + \gamma(t)  \log\frac{D_\theta^t(\mathbf{x}_t,\tilde{\mathbf{y}})}{1-D_\theta^t(\mathbf{x}_t,\tilde{\mathbf{y}})}, \label{eq:new_sampling_eq}\\
    \gamma(t) &= \begin{cases} 
    \gamma & \text{if } t \in (S_{\text{clip\_min}}, S_{\text{clip\_max}}] \\
    0 & \text{otherwise}.
    \end{cases} \nonumber
\end{align}
The gate $\gamma(\cdot)$ enforces that $D$ is used at $S_{\text{clip\_min}} \leq t\leq S_{\text{clip\_max}}$.
This guidance technique is similar to Classifier Guidance proposed by  \citet{dhariwal2021diffusion}.
In practice, every noise schedule is discretized into $N$-step sampling, where $N\leq T$.
Hence, the interval is rounded to the closest sampling steps between $S_{\text{clip\_min}}$ and $S_{\text{clip\_max}}$.
Our experimental results are conducted on limited guidance, showing the trade-off between quality and inference efficiency. The optimal hyper-parameters are searched through FID evaluation and remain consistent across a wide range. 
The details are shown in \cref{subsec:ablation}.

\subsection{Improved Technique for Training Efficiency}
\label{subsec:mixup}

\textbf{Training Pipeline.} \Cref{fig:pipeline,alg:dg-train} illustrate the overall training pipeline for the discriminator.
Initially, the noisy training data $\tilde{\mathcal{D}}$ are filtered into clean subset $\hat{\mathcal{D}}_r$ and corrupt subset $\hat{\mathcal{D}}_f$ with a noise detector (noise is imperfectly filtered).
Then, we augment both training subsets with two proposed data augmentations before feeding it to the discriminator.

\noindent\textbf{Pseudo-clean Shuffle.} 
In practice, unknown and subtle noise rate leads to a high possibility of data imbalance \cite{xia2021sample}. 
In a low noise rate setting, we enrich the corrupt subset by simply swapping the label of clean samples and flip $r$ from $1$ to $0$ accordingly during training.
Using this simple augmentation, $|\hat{\mathcal{D}}_f|$ is scaled proportionally to the label space $K$.
Furthermore, this augmentation improves the robustness of the discriminator to unseen noise in the corrupt set.
Essentially, \textit{it breaks the instance-dependency of several noise types} (such as Asymmetric noise).

\noindent\textbf{SiMix.}
Since the discriminator guided the score direction to avoid the corrupt distribution, the diversity of the generation might deteriorate. 
We propose \textbf{SiMix} (\underline{Si}milarity based \underline{Mix}Up), a new inter-class augmentation inspired by MixUp \cite{zhang2017mixup}, to promote the diversity of corrupt data. 
Given (instance, label) pairs, we combine instances with high structural similarity in the feature space instead of random selection. 
SiMix stabilizes the learning process while preventing the discriminator's gradient from exploding, as $\log \text{D}(\cdot)$ goes up significantly when $\text{D}(\cdot)\to0$.
The detail is shown in \cref{alg:simix}.


\section{Experiments}

\subsection{Experiment Setup}
\hspace{0.9em} To evaluate our method, we conduct experiments with synthetic noise on CIFAR-10, CIFAR-100 \cite{krizhevsky2009learning}, and Tiny-Imagenet \cite{le2015tiny}, including Symmetric \cite{kim2019nlnl}, Asymmetric \cite{han2018co}, and Instance-dependent 
\cite{xia2020part} noises at various scales. 
We synthesize CIFAR-10 noisy data following \citet{bae2022noisy}, and use synthesized CIFAR-100 from TDSM \cite{na2024labelnoise}.
Tiny-Imagenet is perturbed with 20\% Symmetric noise, and the results are reported in \cref{subsec:real_world}.
Synthetic noise details are described in \cref{sec:synthetic_label}.
We also experiment on human-annotated noise on real-world datasets, including FOOD101 \cite{bossard2014food} and Clothing-1M \cite{xiao2015learning}.

\noindent\textbf{Noise Detectors.}
We use CORES \cite{cheng2021learninginstancedependentlabelnoise} and CL \cite{northcutt2021confident} to explore the effect of different noise detection methods on discriminator performance.
The implementation details and results of CORES and CL on each setting are reported in \cref{subsec:noise-detect-imp}.

\noindent\textbf{Evaluation metrics.} We use Inception Score (IS) \cite{NIPS2016_8a3363ab, barratt2018note} and Frechet Inception Distance (FID) \cite{heusel2017gans} to evaluate the sampling quality by comparing distributions of generated and real images. To assess fidelity and diversity, we apply Density (Den.) and Coverage (Cov.) \cite{naeem2020reliable}. Since class-wise insights are essential, we adopt CW-FID, CW-Density (CW-Den.), and CW-Coverage (CW-Cov.) following \cite{chao2022denoising,na2024labelnoise}.

\noindent\textbf{Baselines.} We compare our method with the original model (EDM) training with \cref{eq:noisy_obj} and the state-of-the-art method (TDSM) to demonstrate the effectiveness of our model. 
Furthermore, we retrain the score network on filtered clean data and denote it as "Oracle".
We also compare retraining relabeled datasets from noise detection methods.
GAN-based methods \cite{thekumparampil2018robustness,kaneko2019label} are not reported here because they are prone to mode collapse, and we were unsuccessful in reproducing their experimental results.

\noindent\textbf{Model Configuration.} The score model architecture is the same as the one used in EDM \cite{karras2022elucidating}, while the discriminator architecture is modeled the same as the classifier in TDSM except for the logit layer. 

\noindent\textbf{Data Augmentation.} For \text{Pseudo-clean Shuffle}, based on the noise rate estimated by the noise detector, we apply it to the data with 20\% asymmetric noise and 20\% IDN noise at a random shuffle rate of 30\% to balance the clean and corrupt distribution.
For $\text{SiMix}$, $\lambda$ is sampled from $\text{Beta}(\alpha, \alpha)$. 
In all our experiments, we set $\alpha$ to 0.2 and applied SiMix on half of the mini-batch, while the remaining data remained unchanged.
We used CLIP \cite{radford2021learning} image encoder ViT-B/32 as the feature extractor for $\text{Encode}(\cdot)$ during our experiments.

\begin{table}[t!]
\setlength{\tabcolsep}{1.2pt} 
\centering
\small
\begin{tabular}{lp{3mm}@{\hspace{2mm}} ccc p{0.005mm} ccc c}
\hline
\multirow{2}{*}{CIFAR-100} &  & EDM & TDSM & SBDC & & EDM & TDSM & SBDC & EDM \\
& & ($\dagger$) & (*) & (Ours) & & ($\dagger$) & (*) & (Ours) & ($\dagger$)
\\
\hline
\multirow{2}{*}{Metrics} &  & \multicolumn{3}{c }{Symmetric} & & \multicolumn{3}{c }{Symmetric} & \multirow{2}{*}{Clean} \\ 
\cline{3-5} \cline{7-9} 
\noalign{\vskip 2pt}
& & \multicolumn{3}{c}{20\%} & & \multicolumn{3}{c}{40\%} \\
\hline
\noalign{\vskip 2pt}
FID & ($\downarrow$) & \textbf{2.81} & 4.18 & \underline{3.55} & & \textbf{3.23} & 6.84 & \underline{3.64} & 2.35 \\
IS & ($\uparrow$) & 12.15 & \underline{12.19} & \textbf{12.64} & & 11.83 & \underline{11.96} & \textbf{12.36} & 12.80 \\
Density & ($\uparrow$) & 85.24 & \underline{88.54} & \textbf{98.91} & & 85.00 & \underline{90.05} & \textbf{95.16} & 90.60 \\
Coverage & ($\uparrow$) & \underline{77.71} & 76.99 & \textbf{79.42} & & \underline{76.64} & 73.92 & \textbf{78.51} & 80.42 \\
CW-FID & ($\downarrow$) & 77.64 & \underline{76.77} & \textbf{68.94} & & 97.96 & \underline{91.13} & \textbf{77.13} & 64.66 \\
CW-Den. & ($\uparrow$) & 68.20 & \underline{72.21} & \textbf{90.48} & & 51.39 & \underline{61.27} & \textbf{76.76} & 85.29 \\
CW-Cov. & ($\uparrow$) & 72.13 & \underline{72.36} & \textbf{76.61} & & 63.26 & \underline{65.50} & \textbf{72.03} & 78.75 \\
\hline
\end{tabular}
\caption{Performance comparison on CIFAR-100 dataset under various noise conditions. The percentage below noise type represents the noise rate. The best results are in \textbf{bold} and the second best ones are in \underline{underlined}. $^*$ indicates that we take the results from their checkpoints.} 
\label{tab:cifar100_results}
\end{table}
\subsection{Results On CIFAR-10 And CIFAR-100}

\hspace{0.9em} \Cref{tab:cifar10_var_results,tab:cifar100_results} show the performance of SBDC on CIFAR-10 and CIFAR-100 across various noise settings.
The results indicate SBDC outperforms the state-of-the-art method across class-wise metrics while remaining competitive in the remaining metrics shown in \cref{tab:cifar10_var_results}.
The results exhibit relatively small deviations, ensuring consistency between experiments.
Crucially, SBDC offers a strong performance against small noise settings.
We show in \cref{subsec:ablation} that the significant improvement mostly contributed to the proposed augmentation methods.

The experiments in \cref{tab:cifar100_results} show that our method outperforms the state-of-the-art in all metrics, with FID increasing only 26.33\% compared to 48.75\% of TDSM for the baseline of 20\% noise ratio and 12.69\% compared to 111.76\% for the baseline of 40\% noise ratio. 
The modest improvement of TDSM is likely due to approximating the objective function with a much smaller number of classes, leading to deviations in calculating the actual objective. 
This demonstrates that our method provides a better solution for datasets with a large number of classes.

\Cref{tab:cifar10_compare_oracle} shows that our method delivers superior quality and diversity compared to "Oracle" training and relabeling.
While "Oracle" training reduces noisy labels, it may discard clean samples, limiting diversity. 
Relabeling leverages noisy data but depends on the relabeling method. 
In contrast, our approach directly utilizes noisy data via augmentation while mitigating false generations.

\subsection{Results on Original Annotated Data}
\begin{table}[t!]
\setlength{\tabcolsep}{1.1pt} 
\centering
\small
\begin{tabular}{lc cc p{0.5mm} cc p{0.5mm} ccc}
\hline
\multicolumn{2}{c }{CIFAR-10} & Oracle & Ours & & Oracle & Ours &  & Relabel & Oracle & Ours  
\\
\hline
\multirow{2}{*}{Metric} &  & 
\multicolumn{2}{c }{Clean} & &
\multicolumn{2}{c }{Symmetric} & & \multicolumn{3}{c }{Symmetric} \\ 
\cline{3-4} \cline{6-7} \cline{9-11} 
& & \multicolumn{2}{c}{1.5\%} & &
\multicolumn{2}{c}{20\%} & & 
\multicolumn{3}{c}{40\%} \\
\hline
FID & ($\downarrow$) & \textbf{1.88} & 3.19 &  & \textbf{2.01} & 2.54 &  & 2.83 & \textbf{2.17} & 2.49  \\
IS & ($\uparrow$) & 9.98 & \textbf{10.06} &  & 9.91 & \textbf{10.01} & & 9.70 & 9.76 & \textbf{9.88} \\
Den. & ($\uparrow$) & 105.24 & \textbf{124.40} & & 105.82 & \textbf{116.38} &  & 105.00 & 105.54 & \textbf{116.24} \\
Cov. & ($\uparrow$) &  84.10 & \textbf{85.23} & & 83.76 & \textbf{83.96} & & 82.12 & 83.49 & \textbf{83.94}  \\
\hline
\end{tabular}
\caption{Results on the original CIFAR-10, noisy CIFAR-10 with Relabeled training, and noisy CIFAR-10 with Oracle training. } 
\label{tab:cifar10_compare_oracle}
\end{table}

\begin{figure}[h!]
    \centering
    \includegraphics[width=\linewidth]{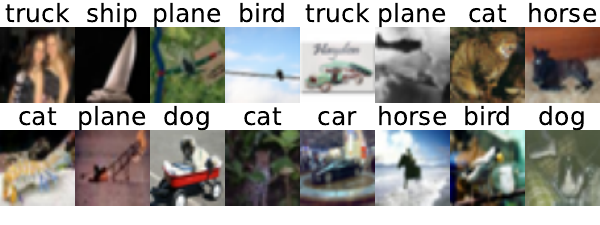}
    \caption{Ambiguous samples of CIFAR-10, filtered by noise detector. The title above images denote ground-truth labels.}
    \label{fig:bad-image}
\end{figure}
To explore potentially noisy or ambiguous labels in the original datasets, we apply CORES to automatically detect it.
From 1.5\% of noisy classified samples, we predict the confidence of their label prediction and visualize 16 images with the lowest confidence in \cref{fig:bad-image}.
It shows that original data also contains examples with noisy or ambiguous labels.
We either apply SBDC or "Oracle" training on them. 
\Cref{tab:cifar10_compare_oracle} shows improved results, indicating that correcting or filtering noisy data enhances learning. 

\subsection{Results On Real World Dataset}
\hspace{0.9em} After filtering clean data (see \cref{subsec:real_world}), we train a ResNet50-based discriminator at 64× resolution for 150 epochs, evaluating every 50 epochs and reporting the best results. 
\Cref{tab:real_results} shows our method outperforms TDSM on most metrics—TDSM struggles on the multi-class FOOD101 dataset, while our approach remains consistently effective, with SBDC yielding superior FID scores.
\begin{table}[t!]
\centering
\setlength{\tabcolsep}{3pt} 
\small
\begin{tabular}{lc ccc p{0.5mm} ccc}
\hline
 &  & EDM & TDSM & SBDC & & EDM & TDSM & SBDC \\
\hline
Metrics &  & \multicolumn{3}{c }{CLOTHING1M} & & \multicolumn{3}{c }{FOOD101}\\ 
\hline
\noalign{\vskip 2pt}
FID & ($\downarrow$) & \underline{4.34} & 6.75 & \textbf{4.17} & & \textbf{3.74}	& 5.21	& \underline{4.41}  \\
IS & ($\uparrow$) & \underline{3.68} & 3.66 & \textbf{3.70} & & 3.93 & \textbf{4.39} & \underline{4.05}  \\
Density & ($\uparrow$) & 98.97 & \textbf{115.8} & \underline{101.3} & & \underline{121.4} & 113.48 & \textbf{127.4}  \\
Coverage & ($\uparrow$) & \underline{74.80} & 74.39	& \textbf{75.53} & & \underline{76.80} & 75.84 & \textbf{76.94} \\
\hline
\end{tabular}
\caption{Performance comparison on CLOTHING1M and FOOD101 datasets. We retrain the baseline and TDSM from scratch with their settings.}
\label{tab:real_results}
\end{table}

\subsection{Ablation Study}
\label{subsec:ablation}
\begin{table}[t!]
\setlength{\tabcolsep}{2.8pt} 
\centering
\small
\begin{tabular}{lccc|cccc}
\hline
\noalign{\vskip 2pt}
\multirow{2}{*}{Metrics} &  P    & R    & F1   & Den.      & Cov.   & CW-Den.    & CW-Cov.   \\ 
\noalign{\vskip 1pt}
\cline{2-8}
\noalign{\vskip 2pt}
&  ($\uparrow$)    & ($\uparrow$)    & ($\uparrow$)   & ($\uparrow$)      & ($\uparrow$)    & ($\uparrow$)    & ($\uparrow$)   \\ 
\noalign{\vskip 1pt}
\hline
\noalign{\vskip 2pt}
CORES.  &  \textbf{87.8} & 58.5 & \textbf{70.2} & 107.4 & 83.7 & 104.5 & 82.7 \\ 
CL &  51.1 & \textbf{83.1} & 63.3 & \textbf{110.5} & \textbf{84.0}  & \textbf{108.3} & \textbf{83.4} \\ 
\noalign{\vskip 1pt}
\hline
\end{tabular}
\caption{Performance comparison between CORES and CL without using SiMix on CIFAR-10 with Asymmetric noise.}
\label{tab:compare_noisy_detect}
\end{table}
\textbf{Noise Detection Impact.} 
We compare CORES and CL on asymmetric noise, as different noise prediction methods impact performance. 
Precision (P) measures accuracy, Recall (R) reflects detected noisy samples, and F1 evaluates overall performance. 
\Cref{tab:compare_noisy_detect} shows that higher Recall improves SBDC, emphasizing the importance of detecting more noisy data over accuracy. This underscores the need for better noise prediction methods.

\begin{figure}[t!]
    \centering
    \begin{subfigure}[b]{0.23\textwidth}
        \centering
        \includegraphics[width=\textwidth]{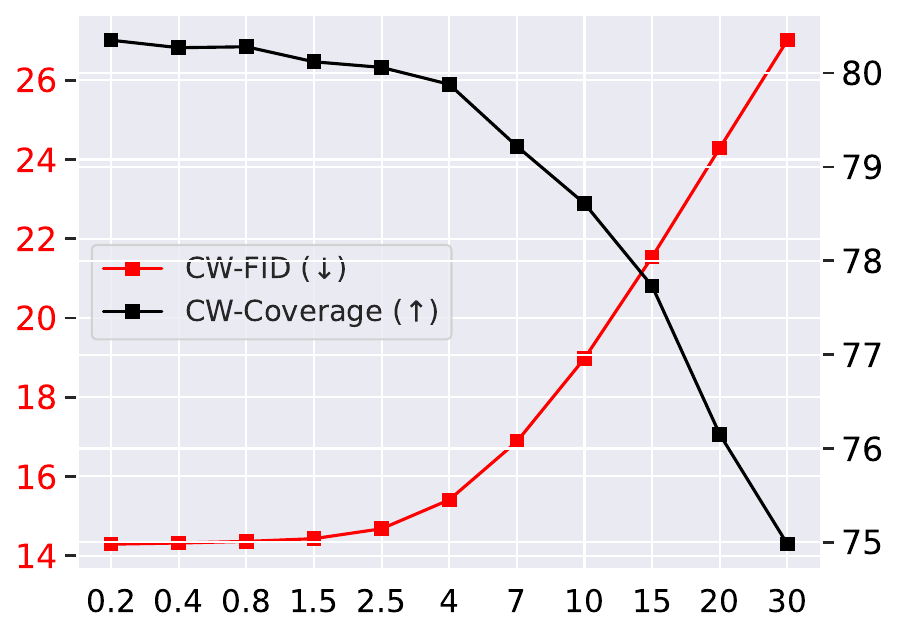}
        \caption{$S_{\text{clip\_min}}$}
        \label{subfig:s_min}
    \end{subfigure}
    \begin{subfigure}[b]{0.23\textwidth}
        \centering
        \includegraphics[width=\textwidth]{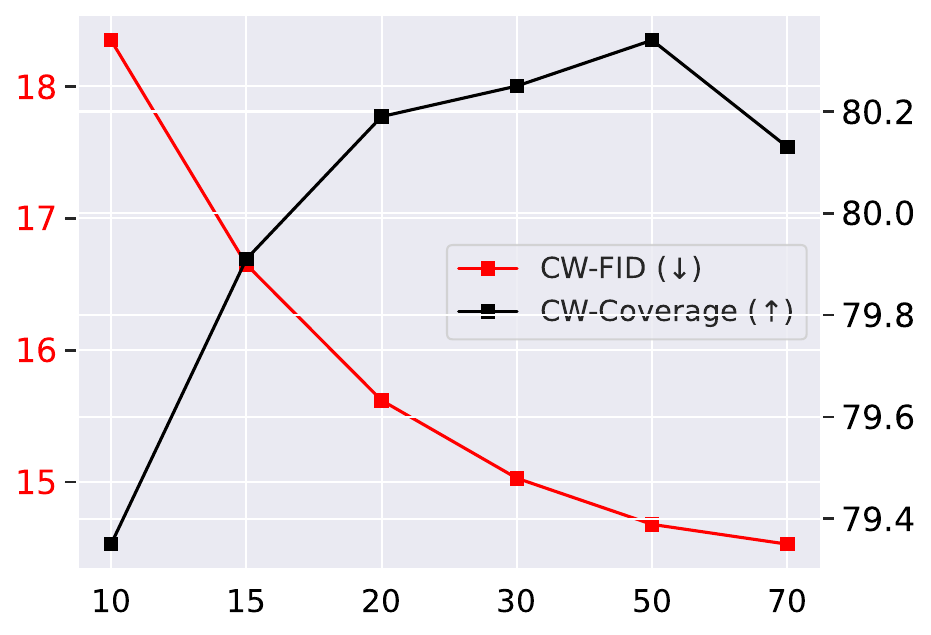}
        \caption{$S_{\text{clip\_max}}$}
        \label{subfig:s_max}
    \end{subfigure}
    \begin{subfigure}[b]{0.27\textwidth}
        \centering
        \includegraphics[width=\textwidth]{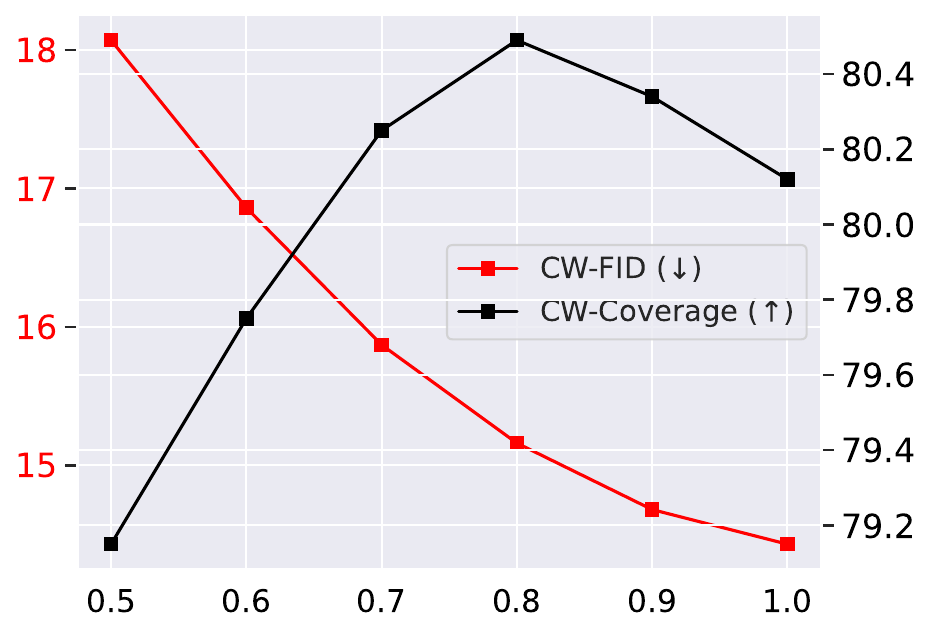}
        \caption{$\gamma$}
        \label{subfig:gamma}
    \end{subfigure}
    \caption{Ablation studies for several parameters, termination point $S_{\text{clip\_min}}$, starting point $S_{\text{clip\_max}}$, guidance strength $\gamma$.}
    \label{fig:abl_hyper}
\end{figure}

\noindent\textbf{Limited Interval Guidance.}
\Cref{subfig:s_min,subfig:s_max} show the changes in CW-FID and CW-Coverage according to $S_{\text{clip\_min}}$ and $S_{\text{clip\_max}}$. 
The results indicate that SBDC has a negligible impact even at a large $S_{\text{clip\_min}}$, while $S_{\text{clip\_max}}$ demonstrates a clear sweet-spot for optimal quality.
Hence, we set $S_{\text{clip\_min}}=1.5$ and $S_{\text{clip\_min}}=50.0$, corresponding to the guidance applied from Step 8 to 16 of the 18 steps for all experiments, including real-world experiments.
Furthermore, we discuss the extension of this wise hyper-parameter searches in other pre-trained DMs in \cref{subsec:extension}.
Some visualization on the intermediate phase of generated samples with/without guidance is shown in \cref{subsec:visualize}.

\noindent\textbf{Guidance Strength.}
\Cref{subfig:gamma} show the optimal $\gamma$ search. A high $\gamma$ value likely leads to a mode drop, similar to Classifier Guidance. Hence, we set $\gamma=0.9$ for balancing between diversity and quality during our all experiments, including real-world experiments.

\noindent\textbf{Data Augmentation.}
As shown in \cref{tab:ablation_augment}, SiMix offers better intra-class quality, likely because the artifact is avoided when mixing instances together. 
We show this through a small example, where mixing two images \(\bigl(\mathbf{x}_1, \mathbf{y}\bigr)\) and \(\bigl(\mathbf{x}_2, \mathbf{y}\bigr)\) from the same class (\(r_1=1,\,r_2=1\)) yields 
\begin{align}
r &= \lambda r_1 + (1-\lambda)r_2 = 1,
\\
L(\mathbf{x}_1, \mathbf{x}_2, \mathbf{y}) 
&= -\log D(\lambda \mathbf{x}_1+(1-\lambda)\mathbf{x}_2, \mathbf{y}),
\end{align}
forcing the discriminator to treat these mixes as normal data. 
However, if the two structures differ significantly, it may produce unnatural samples. 
We compare the performance during the training process with MixUp and SiMix in \cref{subsec:simix_mixup}.
When combined with Pseudo-clean Shuffle, the performance is boosted by a large gap, indicating the model is robust to unseen noise and can even enhance the quality of a well-generated sample. 
Additionally, Pseudo-clean Shuffle stabilizes the learning process of 20\% Asymmetric noise setting, which is often a challenging setting for the noise detector.
\begin{table}[t!]
    \centering
    \begin{tabular}{c c | c c}
        \toprule
        {\textbf{SiMix}} &{\textbf{Pseudo Shuffle}}   & CW-Den. & CW-Cov. \\
        \midrule
         &  & 102.95 & 82.70 \\
         $\checkmark$ & &  104.49 & 82.76 \\
        $\checkmark$ & $\checkmark$ & \textbf{114.80} & \textbf{83.10} \\
        \bottomrule
    \end{tabular}
    \caption{Ablation of proposed data augmentation. Results are reported on CIFAR-10 20\% Asymmetric noise.}
    \label{tab:ablation_augment}
\end{table}

\begin{figure}[t!]
    \centering
    \includegraphics[width=0.5\linewidth]{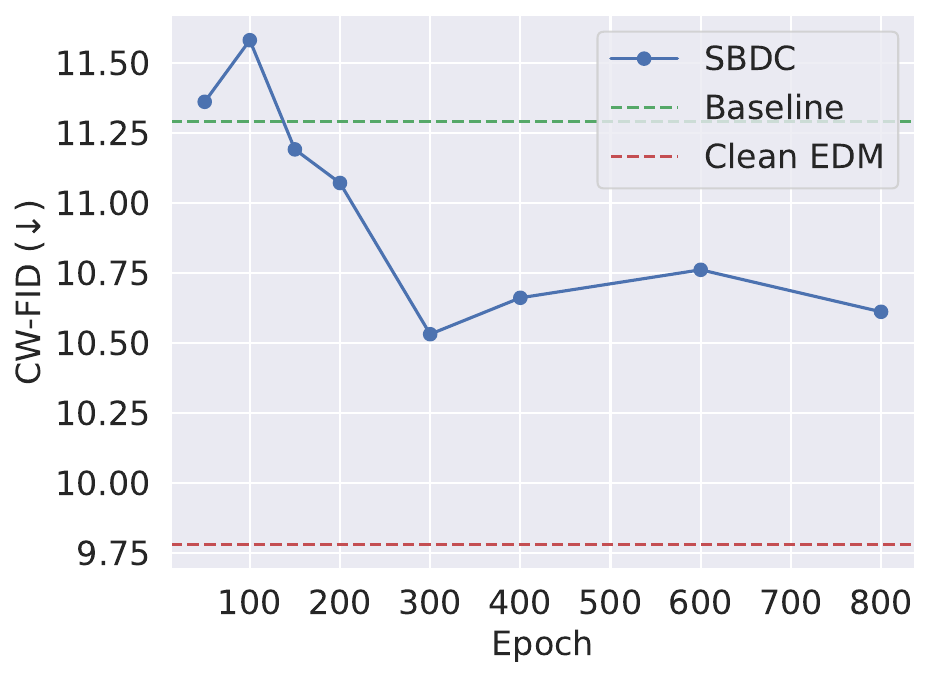}
    \vspace{-2mm}
    \caption{SBDC with different checkpoints of the discriminator.}
    \label{fig:disc_training}
\end{figure}

\noindent\textbf{Discriminator Training.}
Training SBDC typically takes 50-200 epochs, while the classifier in TDSM is 4000 epochs.
\Cref{fig:disc_training} displays the CW-FID curve during discriminator training. The model quickly outperforms the baseline and stabilizes.

\begin{figure}[t!]
    \centering
    \includegraphics[width=0.47\textwidth]{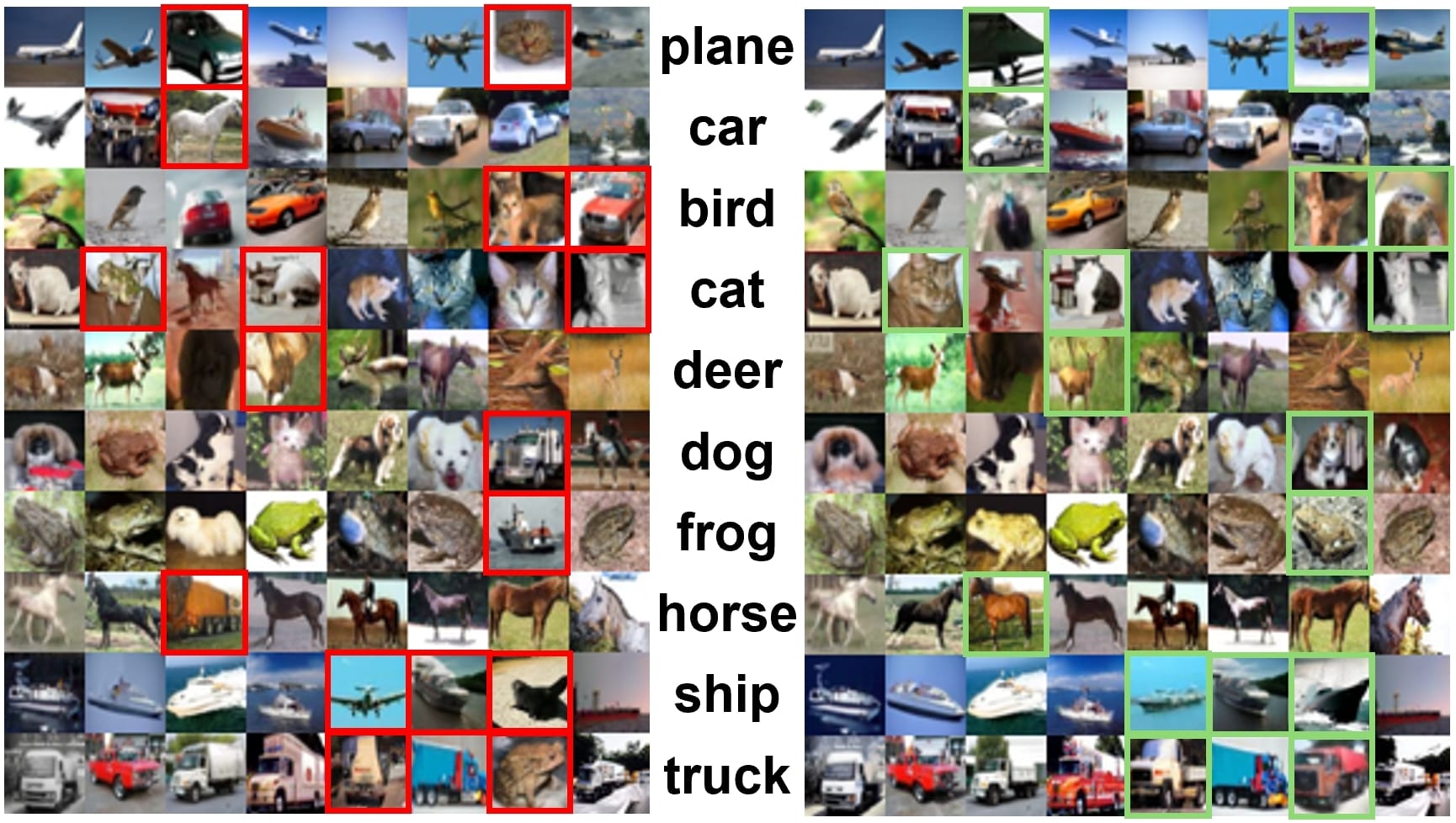}
    \caption{The uncurated generated images of TDSM (left) and TDSM with SBDC (right) on the CIFAR-10 dataset with 50\% symmetric noise. We highlight some error images and corrections in red and green squares, respectively.}
    \vspace{-5mm}
    \label{fig:tdsm-fig}
\end{figure}

Since our method functions as a plug-and-play module, it can also be integrated with methods like TDSM.
We show some generated images from TDSM with/without SBDC in \cref{fig:tdsm-fig}.
The experimental results for applying SBDC to TDSM are described in \cref{subsec:tdsm_sbdc}.

\begin{table}[t!]
\centering
\small
\begin{tabular}{lcc}

\hline
Wall-clock Time & EDM & SBDC \\
\hline
\noalign{\vskip 2pt}
CIFAR-10 & 8m35s & 8m39s \\
CLOTHING-1M & 23m7s & 24m16s \\
\hline
\end{tabular}
\caption{Wall-clock inference time (s) per 50k images on EDM (35 steps) with and without SBDC. Guidance is applied on half of the sampling process.} 
\label{tab:infer_time}
\end{table}

\noindent\textbf{Inference time comparison.} 
All the numbers are measured on 4 NVIDIA A100 GPUs.
In Table~\ref{tab:infer_time}, we report the wall-clock inference time per 100K images on EDM and EDM with SBDC for CIFAR-10 and CLOTHING1M, noting that the inference time of TDSM is the same as EDM.
SBDC does not raise any significant delay because the guidance is applied on only half of the diffusion steps.

\section{Related Works}
\subsection{Supervised Learning} 
\hspace{0.9em} In classification tasks, labels are the learning target and directly impact the models' behaviors.
Based on learning behaviors, researchers have developed methods for improving models' robustness.
\citet{zhang2021understanding,patel2021memorization} showed that the model eventually memorizes noisy data given sufficient training time, however, the model typically remembers clean data before remembering noisy data.
The observation leads to techniques such as early stopping \cite{xia2020robust, liu2020early,bai2021understanding}, curriculum learning \cite{zhou2020robust}, ensemble methods \cite{lee2019robust,han2018co}.
Additionally, noisy models suffer from reduced interpretability, leading to the uncurated representation of data in the latent space. 
\citet{zhu2022detecting,li2022neighborhood,chen2024label} utilized the representations from pre-trained models to mitigate the impact of noise.
\citet{yi2022learning,cheng2021mitigating,kim2024learning} introduced regularization techniques to make the learned representation more robust to noise.
The generalizability of algorithms for learning with noisy labels is also a question that has been actively studied \cite{foret2020sharpness,yu2019does,harutyunyan2020improving}.
In our setting, we aim to learn a model end-to-end without the data pre-processing. 

\subsection{Conditional Generative Learning} 
\hspace{0.9em} Several methods modified the GAN architecture \cite{NIPS2014_5ca3e9b1} to better utilize the label information.
\citet{thekumparampil2018robustness} proposes two methods: one adds a calibration objective to reduce mismatch between generated and true labels, and the other boosts confidence in label predictions. 
Meanwhile, \citet{kaneko2019label} integrates a transition probability matrix into the discriminator, enabling implicit inference of the clean label distribution.

Recently, \citet{na2024labelnoise} integrated transition probability methods into conditional diffusion models, representing a noisy score network as a mixture of clean ones. 
However, their method omits the instance-independent scenario and approximates the score function using only a fraction of classes to reduce training costs—introducing errors that hinder scalability. 
To avoid expensive retraining, we leverage the diffusion model’s sequential generation to directly correct errors, enabling our approach to handle both instance-dependent and instance-independent cases.

\section{Conclusion}
\hspace{0.9em} This paper tackles noisy labels in conditional generative models by analyzing discriminator signals to differentiate between clean and noisy models. 
We validate our approach with two novel data augmentation techniques and highlight its potential for further improvement with advanced noise detection. 
Experiments show our method outperforms others, maintaining efficiency and scalability for large-class datasets.
The method can extend to text condition generative modeling with models like CLIP. 
It would be intriguing to explore zero-shot and few-shot approaches for text noise, especially in privacy and adversarial attacks.
We consider this as a future works.

\newpage

{
    \small
    \bibliographystyle{ieeenat_fullname}

}

\begin{appendix}
\clearpage
\setcounter{page}{1}
\maketitlesupplementary
\section{Proof For Theorem}
\label{sec:proof}

\begin{restatable*}{theorem}{theorem1}
    Let $(\cdot,\mathbf{y}_r)\sim p_r,\ (\cdot,\mathbf{y}_f)\sim p_f$, and $D_\theta^t(\cdot,\cdot)=\sigma(g_\theta(\cdot,\cdot,t))$ be the logistic function of the discriminator. 
    Assume $g$ satisfies the Lipschitz condition: there exists $L>0$ such that for all $t\in[\epsilon,T]$, $\mathbf{x}$, $\mathbf{z}$, and $\mathbf{y}$, we have $||g(\mathbf{x},\mathbf{y},t)-g(\mathbf{z},\mathbf{y},t)||_2^2\leq L||\mathbf{x}-\mathbf{z}||_2^2$. 
    Then, given an optimally trained $D_{\theta^*}$, we have
    \begin{align}
\underset{{\mathbf{x}_t,\mathbf{y},\tilde{\mathbf{y}},\atop \mathbf{y}_r,\mathbf{y}_f}}{\mathbb{E}}&\Big[\Big\|\nabla_{\mathbf{x}_t}\log\frac{D_{\theta^*}(\mathbf{x}_t,\mathbf{y}_r)}{D_{\theta^*}(\mathbf{x}_t,\mathbf{y}_f)}-\nabla_{\mathbf{x}_t}\log\frac{p(\mathbf{x}_t|\mathbf{y})}{p(\mathbf{x}_t|\tilde{\mathbf{y}})}\Big\|_2^2 \Big] \nonumber\\ &\leq L+\underset{{\mathbf{x}_t,\mathbf{y},\tilde{\mathbf{y}}}}{\mathbb{E}} \Big[\big\| \nabla_{\mathbf{x}_t}\log\frac{p(\mathbf{x}_t|\mathbf{y})}{p(\mathbf{x}_t|\tilde{\mathbf{y}})}\big\|_2^2\Big]
    \tag{\ref{eq:dg_ratio}}
\end{align}
    \label{thrm:dg2ratio_app}
\end{restatable*}{theorem}

\noindent\textit{Proof.} 
First, given the objective in \cref{eq:dg_obj}, the optimal discriminator is easily derived as 
\begin{align}
    D_{\theta^*}(\mathbf{x}_t,\mathbf{y}_r) &=\frac{p(\mathbf{x}_t,\mathbf{y}_r)}{p(\mathbf{x}_t,\mathbf{y}_r) + p(\mathbf{x}_t,\mathbf{y}_f)},\\
    D_{\theta^*}(\mathbf{x}_t,\mathbf{y}_f) &=\frac{p(\mathbf{x}_t,\mathbf{y}_f)}{p(\mathbf{x}_t,\mathbf{y}_r) + p(\mathbf{x}_t,\mathbf{y}_f)},
\end{align}
Furthermore, we expose the relationship between $(\mathbf{y}_r,\mathbf{y}_f)$ and $\tilde{\mathbf{y}}$ as,
\begin{align}
    p(\mathbf{y}_r|\mathbf{x}_t)=p(\tilde{\mathbf{y}}, r|&\mathbf{x}_t), \ p(\mathbf{y}_f|\mathbf{x}_t)=p(\tilde{\mathbf{y}}, f|\mathbf{x}_t), \\
    p(\tilde{\mathbf{y}}|\mathbf{x}_t) = p(\tilde{\mathbf{y}}, &r|\mathbf{x}_t) + p(\tilde{\mathbf{y}}, f|\mathbf{x}_t).
\end{align}
Then, the gradient log ratio becomes
\begin{align*}
    \nabla_{\mathbf{x}_t}\log\frac{D_{\theta^*}(\mathbf{x}_t,\mathbf{y}_r)}{D_{\theta^*}(\mathbf{x}_t,\mathbf{y}_f)}&=\nabla_{\mathbf{x}_t}\log\frac{p(\tilde{\mathbf{y}},r|\mathbf{x}_t)}{p(\tilde{\mathbf{y}},f|\mathbf{x}_t)}\nonumber\\&\overset{(i)}{=}\nabla_{\mathbf{x}_t}\log\frac{p_{\theta^*}(r|\mathbf{x}_t,\tilde{\mathbf{y}})}{1-p_{\theta^*}(r|\mathbf{x}_t,\tilde{\mathbf{y}})} \nonumber \\ &
    \overset{(ii)}{=}\nabla_{\mathbf{x}_t}\log\frac{\sigma(g_{\theta^*}(\mathbf{x}_t,\tilde{\mathbf{y}}))}{1-\sigma(g_{\theta^*}(\mathbf{x}_t,\tilde{\mathbf{y}}))}\nonumber\\&=
    \nabla_{\mathbf{x}_t}g_{\theta^*}(\mathbf{x}_t,\tilde{\mathbf{y}}).
\end{align*}
where (i) is due to $p(\cdot,\tilde{\mathbf{y}}|\mathbf{x}_t)=p(\tilde{\mathbf{y}}|\mathbf{x}_t)p_{\theta^*}(\cdot|\tilde{\mathbf{y}},\mathbf{x}_t)$, (ii) is exactly the output of the discriminator, which is a sigmoid function. 
This allows us to break down the LHS in \cref{eq:dg_ratio} to obtain
\begin{align}
    &\text{LHS} \nonumber \\&= \underset{{\mathbf{x}_t,\tilde{\mathbf{y}}}}{\mathbb{E}} \Big[\big\|\nabla_{\mathbf{x}_t}g_\theta(\mathbf{x}_t,\tilde{\mathbf{y}})\big\|_2^2\Big]- 2(F_1(\theta)-F_2(\theta)) + C_1 \nonumber \\ &=
\underset{{\mathbf{x}_t,\tilde{\mathbf{y}}}}{\mathbb{E}} \Big[\big\|\nabla_{\mathbf{x}_t}g_\theta(\mathbf{x}_t,\tilde{\mathbf{y}})\big\|_2^2\Big]- \underset{{\mathbf{x}_t,\tilde{\mathbf{y}}}}{\mathbb{E}} \Big[\big\|\nabla_{\mathbf{x}_t}g_\theta(\mathbf{x}_t,\tilde{\mathbf{y}})\big\|_2^2\Big] \nonumber \\&\quad \ -2F_1(\theta)+2F_2(\theta) + \underset{{\mathbf{x}_t,\tilde{\mathbf{y}}}}{\mathbb{E}} \Big[\big\|\nabla_{\mathbf{x}_t}g_\theta(\mathbf{x}_t,\tilde{\mathbf{y}})\big\|_2^2\Big]+C_1,
\label{eq:LHS}
\end{align}
where
\begin{align*}
    F_1({\theta^*})&=\underset{{\mathbf{x}_t,\mathbf{y},\tilde{\mathbf{y}}}}{\mathbb{E}}[\langle \nabla_{\mathbf{x}_t}g_{\theta^*}(\mathbf{x}_t,\tilde{\mathbf{y}}), \nabla_{\mathbf{x}_t}\log p(\mathbf{x}_t|\mathbf{y})\rangle], \\
    F_2({\theta^*})&=\underset{{\mathbf{x}_t,\mathbf{y},\tilde{\mathbf{y}}}}{\mathbb{E}}[\langle \nabla_{\mathbf{x}_t}g_{\theta^*}(\mathbf{x}_t,\tilde{\mathbf{y}}), \nabla_{\mathbf{x}_t}\log p(\mathbf{x}_t|\tilde{\mathbf{y}})\rangle], \\
    C_1 &= \underset{{\mathbf{x}_t,\mathbf{y},\tilde{\mathbf{y}}}}{\mathbb{E}} \Big[\big\| \nabla_{\mathbf{x}_t}\log\frac{p(\mathbf{x}_t|\mathbf{y})}{p(\mathbf{x}_t|\tilde{\mathbf{y}})}\big\|_2^2\Big] \quad \text{is a constant}.
\end{align*}
\Cref{eq:LHS} can be reformulated in the same manner as \citet{vincent2011connection,song2019generative}, 
which eventually becomes
\begin{align}
    \text{LHS} &= \underset{{\mathbf{x}_0,\mathbf{x}_t,\mathbf{y},\tilde{\mathbf{y}}}}{\mathbb{E}} \Big[\big\|\nabla_{\mathbf{x}_t}g_{\theta^*}(\mathbf{x}_t,\tilde{\mathbf{y}})-\nabla_{\mathbf{x}_t}\log p(\mathbf{x}_t|\mathbf{x}_0,\mathbf{y})\big\|_2^2\Big] \nonumber\\&- \  \underset{{\mathbf{x}_0,\mathbf{x}_t,\tilde{\mathbf{y}}}}{\mathbb{E}} \Big[\big\|\nabla_{\mathbf{x}_t}g_{\theta^*}(\mathbf{x}_t,\tilde{\mathbf{y}})-\nabla_{\mathbf{x}_t}\log p(\mathbf{x}_t|\mathbf{x}_0,\tilde{\mathbf{y}})\big\|_2^2\Big] \nonumber \\ &+ \ \  \underset{{\mathbf{x}_t,\tilde{\mathbf{y}}}}{\mathbb{E}} \Big[\big\|\nabla_{\mathbf{x}_t}g_{\theta^*}(\mathbf{x}_t,\tilde{\mathbf{y}})\big\|_2^2\Big] + C_2 \nonumber \\ &\overset{(i)}{=}
\underset{{\mathbf{x}_0,\mathbf{x}_t,\tilde{\mathbf{y}}}}{\mathbb{E}} \Big[\big\|\nabla_{\mathbf{x}_t}g_{\theta^*}(\mathbf{x}_t,\tilde{\mathbf{y}})-\nabla_{\mathbf{x}_t}\log p(\mathbf{x}_t|\mathbf{x}_0)\big\|_2^2\Big] \nonumber\\&-   \underset{{\mathbf{x}_0,\mathbf{x}_t,\tilde{\mathbf{y}}}}{\mathbb{E}} \Big[\big\|\nabla_{\mathbf{x}_t}g_{\theta^*}(\mathbf{x}_t,\tilde{\mathbf{y}})-\nabla_{\mathbf{x}_t}\log p(\mathbf{x}_t|\mathbf{x}_0)\big\|_2^2\Big] \nonumber \\ &+  \  \underset{{\mathbf{x}_t,\tilde{\mathbf{y}}}}{\mathbb{E}} \Big[\big\|\nabla_{\mathbf{x}_t}g_{\theta^*}(\mathbf{x}_t,\tilde{\mathbf{y}})\big\|_2^2\Big] + C_2 \nonumber \\ &=
\underset{{\mathbf{x}_t,\tilde{\mathbf{y}}}}{\mathbb{E}} \Big[\big\|\nabla_{\mathbf{x}_t}g_{\theta^*}(\mathbf{x}_t,\tilde{\mathbf{y}})\big\|_2^2\Big] + C_2, 
\end{align}
where (i) is the consequence of the unbiased estimator and 
\begin{align}
C_2 &= C_1 + \underset{{\mathbf{x}_0,\mathbf{x}_t,\tilde{\mathbf{y}}}}{\mathbb{E}} \Big[\big\|\nabla_{\mathbf{x}_t}\log p(\mathbf{x}_t|\mathbf{x}_0,\tilde{\mathbf{y}})\big\|_2^2\Big] \nonumber \\ &\quad\quad\ \ - \underset{{\mathbf{x}_0,\mathbf{x}_t,\mathbf{y}}}{\mathbb{E}} \Big[\big\|\nabla_{\mathbf{x}_t}\log p(\mathbf{x}_t|\mathbf{x}_0,\mathbf{y})\big\|_2^2\Big] \nonumber \\ &= C_1
\end{align}
Finally, from our assumption, $g_\theta(\cdot,\mathbf{y},t)$ has Lipschitz constant $L$, then we have
\begin{align}
    \text{LHS} \leq L + C_1
\end{align}

\section{Synthetic Noisy Label Generation Process}
\label{sec:synthetic_label}
\textbf{CIFAR-10} \cite{krizhevsky2009learning} is a $32 \times 32 \times 3$ color image dataset containing 10 classes, with 50,000 training and 10,000 test samples. 
As these datasets are assumed to be noise-free, we introduce three types of noisy labels for label injection. 
Below, we provide details on these noisy label-generation methods. 
Since symmetric noise involves simply assigning a random label, we omit further explanation.

\noindent\textbf{Asymmetric Noise (ASN).} \cite{han2018co,xia2020robust} For this type of noise, we followed the previous works by flipping label classes for CIFAR-10 as shown below.
\begin{table}[h!]
\centering
\small
\begin{tabular}{rcl}
\hline
Truck &$\Rightarrow$& Automobile \\ Bird &$\Rightarrow$ &Airplane \\ Deer& $\Rightarrow$ &Horse \\ Cat &$\Leftrightarrow$ &Dog \\
\hline
\end{tabular}
\label{tab:asym_imp}
\end{table}

\noindent\textbf{Instance Dependent Noise (IDN)} We followed noise generation process as utilized at \citet{xia2020part, cheng2021learninginstancedependentlabelnoise} as shown in \cref{alg:idn_imp}.

\begin{algorithm}[H]
\caption{Instance Dependent Noise Generation Process}
\begin{algorithmic}[1]
\Require Clean samples $(\mathbf{x}_i, \mathbf{y}_i)_{i=1}^n$, Noise rate $\tau$
\State Sample instance flip rates $q \in \mathbb{R}^n$ from the truncated normal distribution $\mathcal{N}(\tau, 0.1^2, [0, 1])$;
\State Independently sample $\mathbf{w}_1, \mathbf{w}_2, ..., \mathbf{w}_c$ from the standard normal distribution $\mathcal{N}(0, 1^2)$;
\For{$i = 1, 2, ..., n$}
    \State $\mathbf{p} = \mathbf{x}_i \times \mathbf{w}_{\mathbf{y}_i};$
    \State $\mathbf{p}_{\mathbf{y}_i} = -\infty;$
    \State $\mathbf{p} = \mathbf{q}_i \times softmax(\mathbf{p});$
    \State $\mathbf{p}_{\mathbf{y}_i} = 1 - \mathbf{q}_i;$
    \State Randomly choose a label from the label space according to the possibilities $\mathbf{p}$ as noisy label $\tilde{\mathbf{y}}_i;$
\EndFor
\Ensure Noisy samples $(\mathbf{x}_i, \tilde{\mathbf{y}}_i)_{i=1}^n$
\end{algorithmic}
\label{alg:idn_imp}
\end{algorithm}

\section{Measure Confidence and Instability}
\label{sec:conf_inst}
\hspace{0.9em} \Cref{alg:conf_inst} directly estimates the confidence in the sampling process instead of the one-step denoise in the forward process.
In our experiments, we use $N = 2,000$ samples to compute the mean $C(t)$ and $I(t)$.

\section{Guidelines for guidance interval searches}

\begin{figure}[t!]
    \centering
    \includegraphics[width=0.64\linewidth]{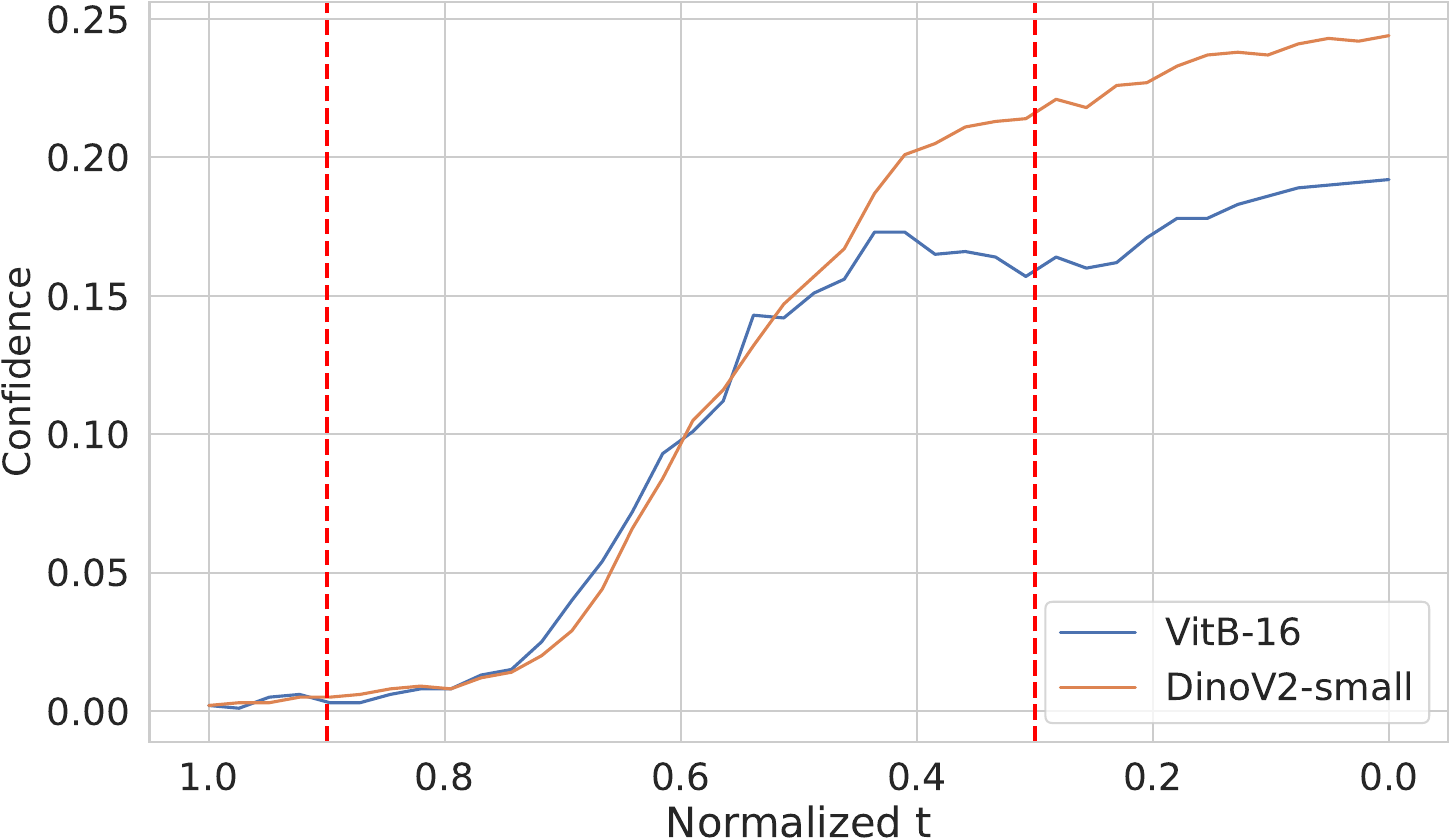}
    \caption{Confidence score on ImageNet 20\% Symmetric noise. The classifiers are selected from open-source \textit{huggingface}.}
    \label{fig:confidence-imagenet}
\end{figure}

\hspace{0.9em} In practice, we determine the interval as follows: 
(1) generate 1,000 samples from the pretrained models, while also storing intermediate denoised samples at each step $t$; 
(2) for each $t$, compute either the \textit{confidence (C)} or \textit{instability (I)} scores (\cref{eq:conf_f,eq:insta_f}); 
(3) finally, plot these values across $t$ and manually select ($S_{\text{clip}\_{\min}}$, $S_{\text{clip}\_{\max}}$) based on the observed trends.
In \cref{fig:confidence-imagenet}, we provide an example using $C$ score on ImageNet, with the chosen $t\in(0.3,45.0)$.
The classifier $f(\cdot)$ choices range from small to large models and are publicly available via \textit{huggingface}. 

\section{Comparison with previous works} 

\hspace{0.9em} \citet{chen2024slight} suggests that a small perturbation may lead to a ``better'' diffusion model.
However, we interpret this to image quality rather than image-condition alignment, as they use FID, IS, Precision, and Recall as standard metrics. 
This might be a capability of classifier-free guidance (CFG), where the ``unconditional term'' can mitigate some label noise. 
However, in \cref{tab:cifar10_metrics}, our findings indicate that CFG remains susceptible to label noise, as evidenced by Classification Accuracy Score (CAS) \cite{ravuri2019classification}, though improved quality.

\begin{table}[h]
\centering
\small
\begin{tabular}{lp{4mm}cccc}
\hline
\textbf{CIFAR-10}  & & \textbf{Clean} & \textbf{Symm.} & \textbf{Asymm.} & \textbf{IDN} \\
\hline
\textbf{Metric} & & \textbf{0\%} & \textbf{20\%} & \textbf{20\%} & \textbf{20\%}                \\
\hline
FID & ($\downarrow$)      & 5.68  & 5.11  & 5.29  & 4.99 \\
IS & ($\uparrow$)       & 3.83  & 4.24  & 3.95  & 4.29 \\
Precision & ($\uparrow$)& 70.53 & 67.28 & 69.40 & 67.63 \\
Recall & ($\uparrow$)   & 50.57 & 55.50 & 52.60 & 55.63 \\
CAS & ($\uparrow$) & \textbf{73.51} & 69.39 & 68.74 & 68.08 \\
\hline
\end{tabular}
\caption{CIFAR-10 trained with CFG, where label noises vary.}
\vspace{-5mm}
\label{tab:cifar10_metrics}
\end{table}

\section{Additional Results}
\hspace{0.9em} In this section, we include more details about the sample selection and sampling of SBDC. All the experiments are run on 4 NVIDIA A100 GPUs.

\begin{figure*}[t!]
    \centering
    \begin{subfigure}[]{0.18\textwidth}
        \centering
        \includegraphics[width=\textwidth]{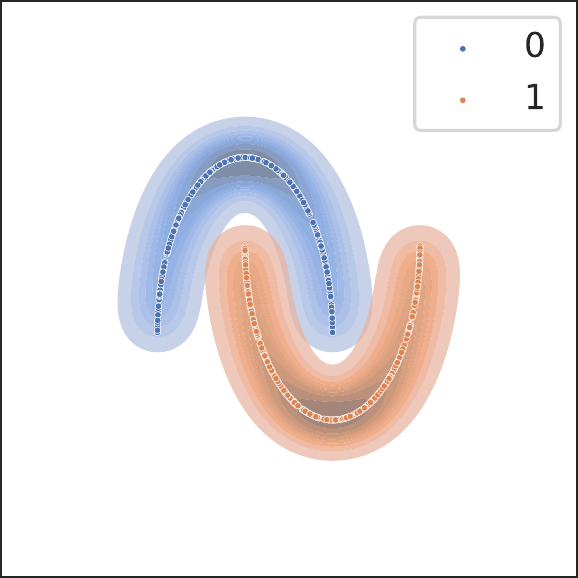}
        \caption{Toy data}
        \label{fig:toy_data}
    \end{subfigure}
    \hfill
    \begin{subfigure}[]{0.18\textwidth}
        \centering
        \begin{subfigure}[]{\textwidth}
            \centering
            \includegraphics[width=\textwidth]{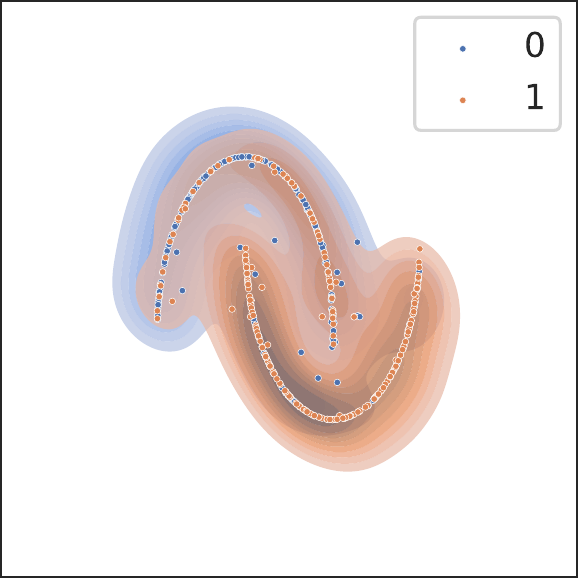}
        \end{subfigure}
        \vspace{0.5em}
        \begin{subfigure}[]{\textwidth}
            \centering
            \includegraphics[width=\textwidth]{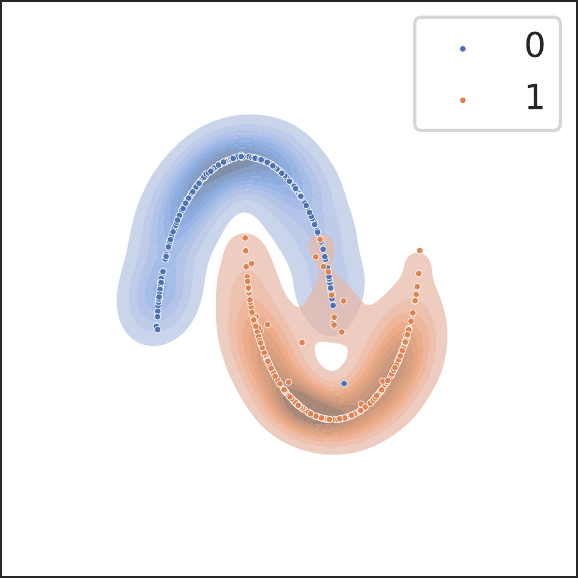}
        \end{subfigure}
        \caption{Generated samples }
        \label{subfig:gen_sample}
    \end{subfigure}
    \hfill
    \begin{subfigure}[]{0.18\textwidth}
        \centering
        \begin{subfigure}[]{\textwidth}
            \centering
            \includegraphics[width=\textwidth]{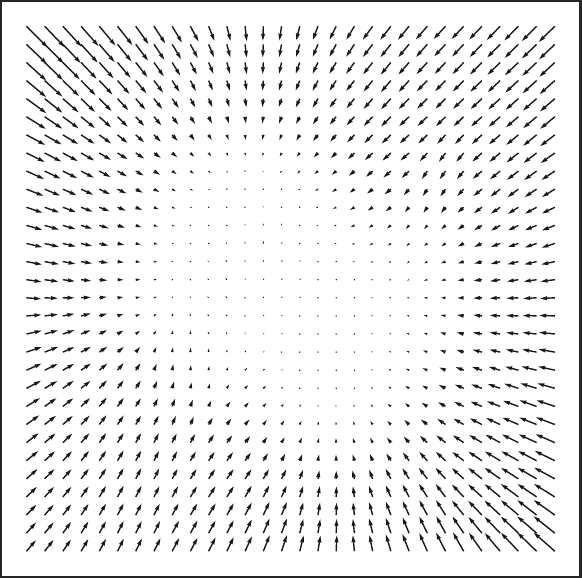}
        \end{subfigure}
        \vspace{0.5em}
        \begin{subfigure}[]{\textwidth}
            \centering
            \includegraphics[width=\textwidth]{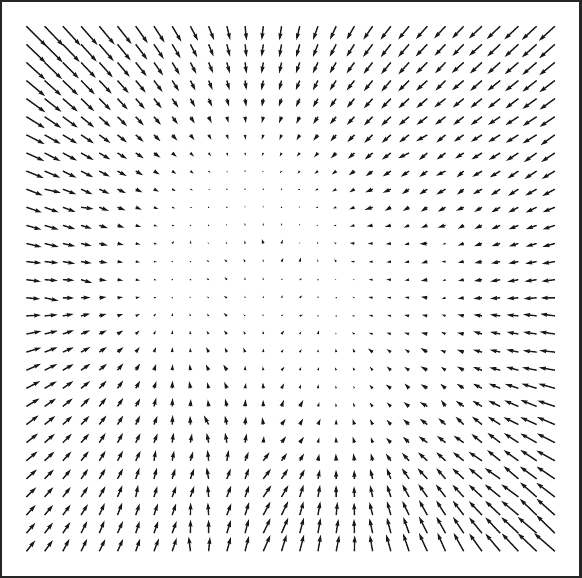}
        \end{subfigure}
        \caption{Class 0 gradient}
        \label{subfig:class_1_grad}
    \end{subfigure}
    \hfill
    \begin{subfigure}[]{0.18\textwidth}
        \centering
        \begin{subfigure}[]{\textwidth}
            \centering
            \includegraphics[width=\textwidth]{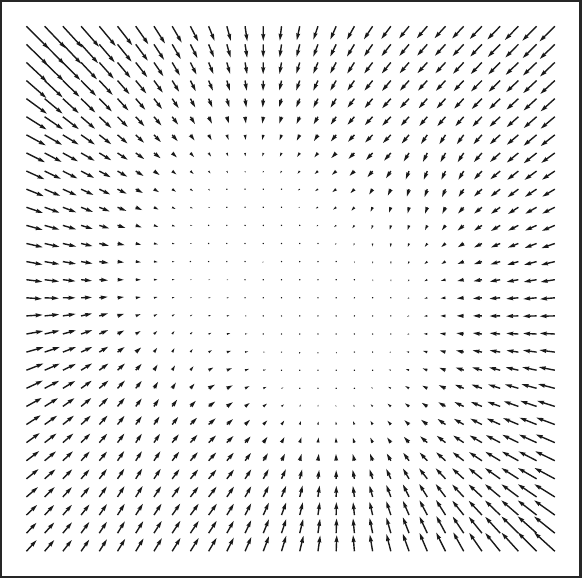}
        \end{subfigure}
        \vspace{0.5em}
        \begin{subfigure}[]{\textwidth}
            \centering
            \includegraphics[width=\textwidth]{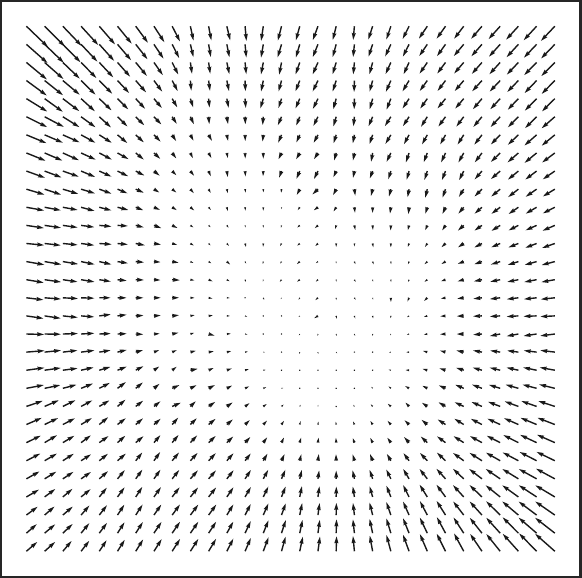}
        \end{subfigure}
        \caption{Class 1 gradient}
        \label{subfig:class_2_grad}
    \end{subfigure}
    \hfill
    \begin{subfigure}[]{0.18\textwidth}
        \centering
        \includegraphics[width=\textwidth]{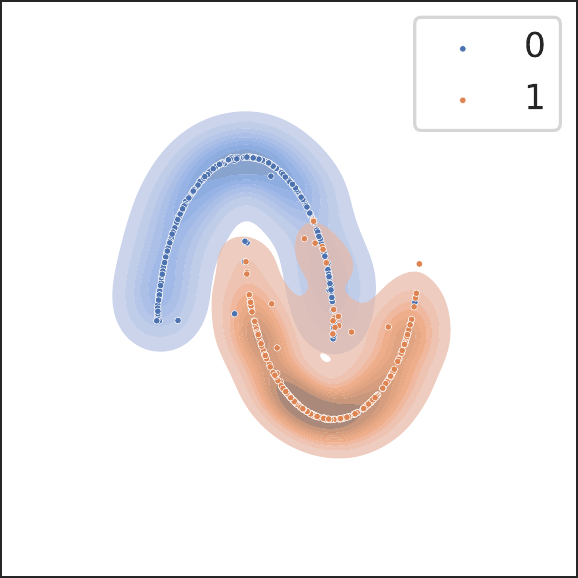}
        \caption{Discriminator training with mixed data}
        \label{subfig:mix_train}
    \end{subfigure}
    \caption{Comparison of the baseline model (first row) with our proposed method (second row). Discriminator Guidance is applied with $\gamma=1.5$. }
    \label{fig:toy_exp}
\end{figure*}

\subsection{Toy Experiments}
\hspace{0.9em} We show our analysis on the inter-twinning moon dataset in \cref{fig:toy_exp}. 
The discriminator network is a 4-layered MLP with 256 neurons, 50\% label of all samples are flipped and we only use half of the training data to train the discriminator network. 
Without proper guidance, the inaccurate score $s_\theta$ produced samples from a noisy distribution, as illustrated in \cref{subfig:gen_sample}. 
In contrast, $s_\theta$ with the discriminator signal effectively steers $s_\theta$ towards $\nabla \log p_r^t$.

In practice, noise detection methods can still make mistakes in distinguishing samples. 
Therefore, we mixed corrupted samples into clean data at a ratio of 1:2 and vice versa for the discriminator training. 
\Cref{subfig:mix_train} shows that SBDC is mostly unaffected by noise in non-overlapping regions, while it may struggle to accurately classify samples in more complex regions. 
However, this can be mitigated by optimizing the parameters of the sampling process.

\subsection{Noise Detection Implementation Details}
\label{subsec:noise-detect-imp}
\begin{table}[h!]
\setlength{\tabcolsep}{3.5pt} 
\centering
\small
\begin{tabular}{lc ccc ccc p{1mm} cc}

\hline
\multirow{2}{*}{} &  & \multicolumn{6}{c }{CIFAR-10} & & \multicolumn{2}{c }{CIFAR-100} \\
\cline{3-8} \cline{10-11}
\noalign{\vskip 2pt}
& & 20 S & 20 A & 20 I & 40 I & 50 S & 80 S & & 20 S & 40 S\\
\hline
\noalign{\vskip 2pt}
P & ($\uparrow$) & 87.1 & 87.8 & 87.7 & 94.7 & 96.2 & 79.8 & & 94.1 & 97.5 \\
R & ($\uparrow$) & 92.80 & \textbf{58.5} & 92.2 & 92.0 & 92.1 & 90.1 & &  84.8 & 82.9 \\
F1 & ($\uparrow$) & 89.84 & 70.2 & 89.9 & 93.3 & 94.1 & 84.6 & & 89.2 & 89.6 \\
\hline
\end{tabular}
\caption{Performance of CORES on CIFAR-10 and CIFAR-100 datasets. S, A, and I denote Symmetric, Asymmetric, and IDN, respectively. Lower Recall undermines the robustness of the discriminator. This encourages prioritizing models capable of capturing more noisy data.} 
\label{tab:cores_perform}
\end{table}

\begin{algorithm}[t!]
\caption{\textbf{Measurement of Confidence \& Instability}}
\label{alg:conf_inst}
\begin{algorithmic}[1]
\State \textbf{Initialize} Pre-trained classifier $f$, $C[t] = list() \ \forall t \in \{1, \dots, T\}$, $I[t] = list()\ \forall t \in \{1, \dots, T-1\}$ 
\Repeat
    \State $\hat{\mathbf{x}}_T \sim \mathcal{N}(0, \mathbf{I})$, $\mathbf{y}\sim[K]$  
    \For{$t = T, \dots, 1$}
        \State Compute $\mathbf{y}_{\text{prev}}^t=f(\mathbf{x}_\theta(\hat{\mathbf{x}}_t,\mathbf{y}))$
        \If{$t \neq T$} 
            \State 
            $I[t].append(\mathbbm{1}[\mathbf{y}_{\text{prev}}^t\neq\mathbf{y}_{\text{prev}}^t])$
            
        \EndIf
        \State $C[t].append(\mathbbm{1}[\mathbf{y}_{\text{prev}}^t=\mathbf{y}])$
        \State Sample $\hat{\mathbf{x}}_{t-1} \sim q(\hat{\mathbf{x}}_{t-1}|\hat{\mathbf{x}}_t,\mathbf{x}_\theta(\hat{\mathbf{x}}_t,\mathbf{y}))
$
    \EndFor
\Until{$N$ iterations}
\For{$t = T, \dots, 1$}
    \State $C(t) = \texttt{numpy.mean}(C[t])$
\EndFor
\For{$t = T-1, \dots, 1$}
    \State $I(t) = \texttt{numpy.mean}(I[t])$
\EndFor
\end{algorithmic}
\end{algorithm}
\noindent\textbf{CORES.} To make the detection process of CORES more effective, we use the feature extracted from the pre-trained CLIP model, specifically the ViT-B/32 version, freeze it, and fine-tune the linear classifier layer. 
We also employ differentiable augmentations \cite{zhao2020differentiable}, containing horizontal flip, vertical flip, isotropic scaling, anisotropic scaling, fractional translation, and fractional rotation to provide the method with more diversity in detections.
We train the model for 100 epochs, with a batch size of 64 on CIFAR-10 and CIFAR-100.
We set the learning rate to 0.1 at the first 55 epochs and then decay it by a factor of 10 for the following training epochs.
Other configurations are kept the same as specified in their paper.
The noisy samples are taken from the last epoch and we report the corrupted samples detection performance of CORES at the last epoch in Table~\ref{tab:cores_perform}. 

It shows the relatively high-quality detection of CORES in different settings. 
However, it remains suffering from asymmetric noise, which results in worse performance.

\noindent\textbf{Confident Learning.} We kept the setting of CL for Asymmetric noise the same as in \citet{pleiss2020identifying}, where the model selected is ResNet-50, training with the learning rate $0.1$ for epoch $[0,150)$, $0.01$ for epoch $[150,250)$, $0.001$ for epoch $[250,350)$, momentum $0.9$, and weight decay $0.0001$.


\subsection{Real-world Experimental Details}
\label{subsec:real_world}
\hspace{0.9em} Food101 \cite{bossard2014food} is a color image dataset containing 101 food categories, with 1,000 samples per category. 
Each class includes 250 images annotated by humans for testing, while the remaining 750 images have real-world label noise. 
Clothing-1M \cite{xiao2015learning} is another real-world dataset with noisy labels, sourced from various online shopping websites, and consists of 1 million training images across 14 classes.
We resize the dataset to resolution $64\times64$ for faster training.
First, in the noise prediction stage, we use the default settings of CORES \cite{cheng2021learninginstancedependentlabelnoise} and SIMIFEAT-r \cite{zhu2022detecting} for CLOTHING1M and FOOD101 respectively. 
Specifically, in each noise detection epoch, we use a batch size of 32 and sample 1000 mini-batches from the training data. 
The training process for CLOTHING1M has 120 epochs while FOOD101 has 100 epochs.

\noindent\textbf{Clean sample selection.} 
We label any sample flagged as noisy over the past 40 epochs as corrupted, except FOOD101, where high complexity leads to many clean samples being misclassified.
We set an upper threshold of 5 to identify clean data.
Both methods use the CLIP ViT-B/32 feature extractor.

\begin{table}[t!]
\centering
\small
\begin{tabular}{lc ccc p{0.5mm} ccc}
\hline
 &  & \multicolumn{3}{c }{Tiny-Imagenet 200} \\
\hline
Metric &  & EDM & TDSM & SBDC \\ 
\hline
\noalign{\vskip 2pt}
FID & ($\downarrow$) & 24.16 & 22.57 & \textbf{21.19}  \\
IS & ($\uparrow$) & 11.01 & 11.61 & \textbf{11.87} \\
Density & ($\uparrow$) & 48.80 & 47.93 & \textbf{55.21}  \\
Coverage & ($\uparrow$) & 28.09 & 28.96	& \textbf{31.06} \\
\hline
\end{tabular}
\caption{Performance comparison on Tiny-Imagenet 200 dataset. We retrain the baseline and TDSM from scratch with their settings.}
\label{tab:tiny_results}
\end{table}
\noindent\textbf{TinyImagenet Experiment.} 
We use CORES to filter the noisy data.
\Cref{tab:tiny_results} shows the proposed method consistently outperforms the baseline in the real-world dataset with a large number of classes.

\subsection{Experiments on ImageNet-128}
We trained the diffusion models on ImageNet-128 with 20\% Symmetric noise for 300K iterations. 
\Cref{tab:imagenet-result} presents the results with full trajectory guidance (SBDC-$f$) and with $\gamma$-gate (SBDC-$\gamma$), where $t\in[0.3, 45.0]$, demonstrating that SBDC remains effective even when scaling up.

\begin{table}[t!]
\setlength{\tabcolsep}{4pt} 
\centering
\small
\begin{tabular}{lcccccc}
\hline
\textbf{ImageNet} & \textbf{FID}$\downarrow$  & \textbf{IS}$\uparrow$   & \textbf{Pre.}$\uparrow$  & \textbf{Rec.}$\uparrow$ & \textbf{Den.}$\uparrow$  & \textbf{Cov.}$\uparrow$ \\
\hline
Original & 30.03 & 23.32 & 55.29 & \textbf{57.99} & 59.14 & 66.16  \\
SBDC-$f$ & \textbf{26.66} & \textbf{27.63} & \textbf{59.68} & 55.65 & \textbf{66.51} & \textbf{71.40} \\
SBDC-$\gamma$ & 26.94 & 27.19 & 59.40 & 55.91 & 66.17 & 71.33 \\
\hline
\end{tabular}
\caption{Comparison on ImageNet 20\% Symm. noise.}
\label{tab:imagenet-result}
\end{table}

\subsection{Sensitivity analysis on different noise detectors.}
In \cref{tab:sensitivity_anal}, we present the effect of different noise detectors on the performance of SBDC.
The results indicate that the performance does not vary significantly across detectors.
In practice, the noise detector can be replaced with a strong pretrained classifier to enhance detection performance.

\begin{table}[t!]
\setlength{\tabcolsep}{3pt} 
\centering
\small
\begin{tabular}{lp{5mm} ccccc}
\hline
\multirow{1}{*}{CIFAR-10} &  & Orig. & CORES & AUM & Simi-v & Simi-r \\
\hline
ND Precision & ($\uparrow$) & -	& 94.7 & 94.18 & 88.75 & 85.30 \\
ND Recall & ($\uparrow$) & - & 92.0 & 86.09 & 91.43 & 92.16\\
\midrule
FID & ($\downarrow$) & 1.98	& 2.49 & 2.57 &	2.34 & 2.51 \\
Density & ($\uparrow$) & 103.24	& 116.24 & 117.41 & 115.13 & 115.09  \\
Coverage & ($\uparrow$) & 83.14 & 83.94 & 83.86 & 83.99 & 83.60  \\
CW-FID & ($\downarrow$) & 29.72 & 13.81 & 14.28 & 13.96 &  14.50  \\
CW-Den. & ($\uparrow$) & 75.39 & 106.77 & 108.07 & 104.96 & 105.32  \\
CW-Cov. & ($\uparrow$) & 73.62 & 81.33 & 81.32 & 81.34 & 81.11  \\
\hline
\end{tabular}
\vspace{-2mm}
\caption{CIFAR-10 40\% IDN results under various ND methods.} 
\vspace{-3mm}
\label{tab:sensitivity_anal}
\end{table}

\begin{table*}[t!]
\centering
\small
\begin{tabular}{lp{5mm} ccc p{0.005mm} ccc p{0.005mm} ccc}
\hline
\multirow{2}{*}{CIFAR-10} &  & TDSM & SBDC & SBDC & & TDSM & SBDC & SBDC & & TDSM & SBDC & SBDC \\
& &  & (EDM) & (TDSM) & &  & (EDM) & (TDSM) & &  & (EDM) & (TDSM)
\\
\hline
\multirow{2}{*}{Metrics} &  & \multicolumn{3}{c}{Symmetric} & & \multicolumn{3}{c}{Symmetric} & & \multicolumn{3}{c}{Symmetric} \\ 
\cline{3-5} \cline{7-9} \cline{11-13}
\noalign{\vskip 2pt}
& & \multicolumn{3}{c}{20\%} & & \multicolumn{3}{c}{50\%} & & \multicolumn{3}{c}{80\%}  \\
\hline
\noalign{\vskip 2pt}
FID & ($\downarrow$) & \textbf{2.16} & 2.54 & 2.95 & & 2.43 & \textbf{2.29} &  3.31 & &  2.22 & \textbf{2.06} &  2.43\\
IS & ($\uparrow$) & \textbf{10.01} & \textbf{10.01} & 9.92 & & 9.81 & \textbf{9.86} & 9.80 & & 9.76 & \textbf{9.83} & 9.78 \\
Density & ($\uparrow$) & 113.34 & 116.38 & \textbf{124.30} & &  113.13 & 114.22 & \textbf{126.05} & & 108.10 & 103.66 & \textbf{109.06} \\
Coverage & ($\uparrow$) & \textbf{84.94} & 83.96 & 84.82 & &  84.13 & 83.85 & \textbf{84.20} & &  \textbf{83.90} & 83.04 & 83.69\\
CW-FID & ($\downarrow$) & \textbf{11.00} & 11.81 & 11.79 & &  18.20 & 16.89 & \textbf{14.47} & & 59.93 & 48.69 & \textbf{37.11} \\
CW-Density & ($\uparrow$) & 108.34 & 112.15 & \textbf{122.04} & &  94.87 & 97.72 & \textbf{115.75} & &  52.29 & 56.72 & \textbf{70.23}\\
CW-Coverage & ($\uparrow$) & 83.66 & 82.60 & \textbf{83.94} & &  79.18 & 79.46 & \textbf{81.54} & & 56.44 & 59.09 & \textbf{67.98}\\
\hline
\rule{0pt}{2pt}
\multirow{2}{*}{Metrics} & & \multicolumn{3}{c}{Asymmetric} & & \multicolumn{3}{c}{Instance} & & \multicolumn{3}{c}{Instance} \\
\cline{3-5} \cline{7-9} \cline{11-13}
\noalign{\vskip 2pt}
& & \multicolumn{3}{c}{20\%} & & \multicolumn{3}{c}{20\%} &  & \multicolumn{3}{c}{40\%} \\
\hline
\noalign{\vskip 2pt}
FID & ($\downarrow$) & 2.38 & \textbf{2.15} & 2.97 & & 2.43 & \textbf{2.42} &  3.45& &  \textbf{2.22} & 2.59 & 3.08\\
IS & ($\uparrow$) & 10.10 & 9.90 & \textbf{10.13} & & 9.73 & \textbf{9.88} & 9.74 & & 9.85 & \textbf{10.09} & 9.93 \\
Density & ($\uparrow$) &  115.81 & 107.73 & \textbf{122.26} & & 113.12 & 114.98 & \textbf{126.06} & & 112.10 & 113.70 & \textbf{122.97} \\
Coverage & ($\uparrow$) & \textbf{85.03} & 83.56 & 84.99 & & \textbf{84.03} & 83.94 & 83.80 & & \textbf{84.51} & 83.64 & 84.25 \\
CW-FID & ($\downarrow$) &  10.94 & \textbf{10.61} & 11.95 & &  12.05 & \textbf{11.63} & 12.78 & & 18.10 & 14.68 & \textbf{13.59}\\
CW-Density & ($\uparrow$) & 113.28 & 104.74 & \textbf{121.10} & & 106.87 & 109.96 & \textbf{123.51} & & 94.19 & 103.30 & \textbf{114.12}\\
CW-Coverage & ($\uparrow$) & 84.32 & 82.86 & \textbf{84.38} & & 82.57 & 82.52 & \textbf{82.86} & & 80.08 & 80.34 & \textbf{82.21}\\
\hline
\end{tabular}
\caption{Performance comparison on CIFAR-10 dataset for different methods. We specify the model in which we apply SBDC in the parenthesis.}
\label{tab:cifar10_sbdc}
\end{table*}

\subsection{Extension to Others Pre-trained Diffusion Models}
\label{subsec:extension}
\begin{table}[h!]
\setlength{\tabcolsep}{1.2pt} 
\centering
\small
\begin{tabular}{lp{3mm}@{\hspace{2mm}} ccc p{0.005mm} ccc}
\hline
CIFAR-10 & & \multicolumn{7}{c }{40\% IDN Noise} \\ 
\hline
\multirow{2}{*}{Metrics} &  & {EDM} & {Ours} & {Clean} & & {EDM} & {Ours} & {Clean} 
\\
\cline{3-5} \cline{7-9} 
\noalign{\vskip 2pt}
& & \multicolumn{3}{c }{NFE=35 ($0.25,0.75$)} & & \multicolumn{3}{c }{NFE=512 ($0.55,0.85$)} \\ 
\hline
\noalign{\vskip 2pt}
FID & ($\downarrow$) & 9.56 & \textbf{8.85} & 9.70 & & 2.22 & \textbf{2.21} & 2.16 \\
IS & ($\uparrow$) & 10.17 & \textbf{10.27} & 10.30 & & 9.85 & \textbf{9.87} & 9.98 \\
Density & ($\uparrow$) & 81.67 & \textbf{95.28} & 83.54 & & 103.66 & \textbf{106.63} & 107.88 \\
Coverage & ($\uparrow$) & 70.98 & \textbf{71.61} & 71.80 & & 83.02 & \textbf{83.73} & 84.26 \\
CW-FID & ($\downarrow$) & 37.68 & \textbf{20.63} & 18.94 & & 29.75 & \textbf{24.91} & 10.05 \\
CW-Den. & ($\uparrow$) & 60.96 & \textbf{93.91} & 82.18 & & 76.20 & \textbf{82.54} & 107.43 \\
CW-Cov. & ($\uparrow$) & 61.43 & \textbf{70.37} & 71.32 & & 74.04 & \textbf{76.56} & 84.16 \\
\hline
\end{tabular}
\caption{Performance comparison on CIFAR-10 dataset with VP noise schedule. 
The number in parenthesis corresponds to $(S_{\text{clip}\_\min},S_{\text{clip}\_\max})$.
The best results are in \textbf{bold}.} 
\label{tab:vp_results}
\end{table}

\hspace{0.9em} In \cref{tab:vp_results}, we show that the proposed method can also be extended to Variance-Preserving (VP) noise schedule \cite{song2020denoising} effectively and it indeed improves performance at different number of sampling steps.

\subsection{TDSM with Score-based Discriminator Correction}
\label{subsec:tdsm_sbdc}
\hspace{0.9em} We conducted experiments by directly applying guidance on TDSM. 
We also report the result on Symmetric noise with 20\% noise rate.
The experimental results in \cref{tab:cifar10_sbdc,tab:cifar100_sbdc} show competitive results against EDM with SBDC. 
\begin{table}[h!]
\setlength{\tabcolsep}{1.2pt} 
\centering
\small
\begin{tabular}{lc ccc p{0.5mm} ccc}
\hline
\multirow{2}{*}{CIFAR-100} &  & TDSM & SBDC & SBDC & & TDSM & SBDC & SBDC \\
& &  & (EDM) & (TDSM) & & & (EDM) & (TDSM)
\\
\hline
\multirow{2}{*}{Metrics} &  & \multicolumn{3}{c }{Symmetric} & & \multicolumn{3}{c }{Symmetric} \\ 
\cline{3-5} \cline{7-9} 
\noalign{\vskip 2pt}
& & \multicolumn{3}{c}{20\%} & & \multicolumn{3}{c}{40\%} \\
\hline
\noalign{\vskip 2pt}
FID & ($\downarrow$) & 4.18 & \textbf{3.55} & 4.65 & & 6.84 & \textbf{3.64} & 6.51 \\
IS & ($\uparrow$) & 12.19 & 12.64 & \textbf{12.68} & & 11.96 & 12.36 & \textbf{12.51} \\
Density & ($\uparrow$) & 88.54 & 98.91 & \textbf{101.67} & & 90.05 & 95.16 & \textbf{99.67} \\
Coverage & ($\uparrow$) &  76.99 & \textbf{79.42} & 78.96 & & 73.92 & \textbf{78.51} & 75.95 \\
CW-FID & ($\downarrow$) & 76.77 & \textbf{68.94} & 69.45 & & 91.13 & \textbf{77.13} & 78.05 \\
CW-Den. & ($\uparrow$) & 72.21 & 90.48 & \textbf{94.04} & & 61.27 & 76.76 & \textbf{82.24} \\
CW-Cov. & ($\uparrow$) & 72.36 & \textbf{76.61} & 76.39 & & 65.50 & \textbf{72.03} & 70.95 \\
\hline
\end{tabular}
\caption{Performance comparison on CIFAR-100 dataset for different methods. We specify the model in which we apply SBDC in the parenthesis.} 
\label{tab:cifar100_sbdc}
\vspace{-3mm}
\end{table}

\subsection{Comparison between SiMix and MixUp}
\label{subsec:simix_mixup}
\begin{figure}[h!]
    \centering
    \begin{subfigure}[b]{0.23\textwidth}
        \centering
        \includegraphics[width=\textwidth]{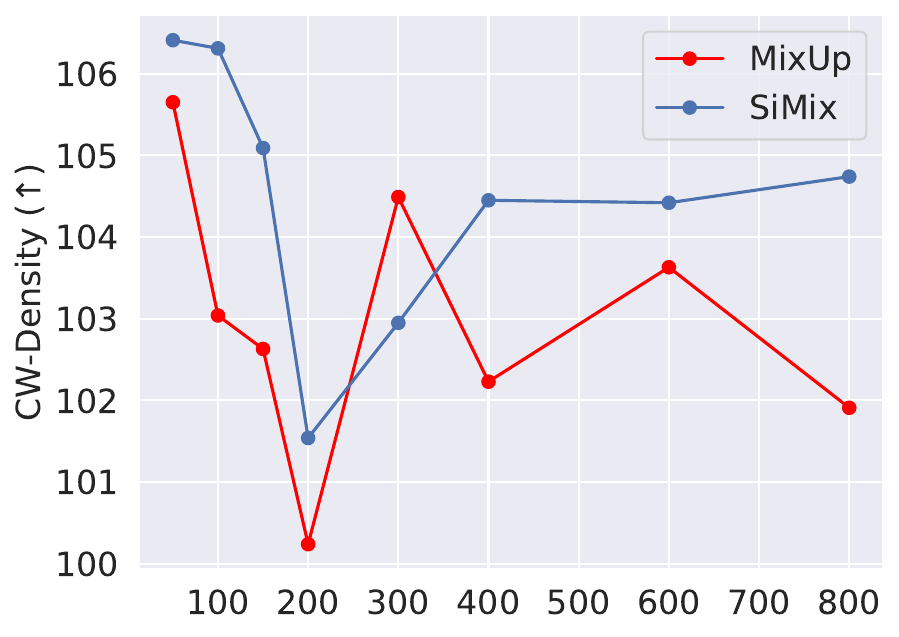}
        \label{subfig:simix_den}
    \end{subfigure}
    \begin{subfigure}[b]{0.23\textwidth}
        \centering
        \includegraphics[width=\textwidth]{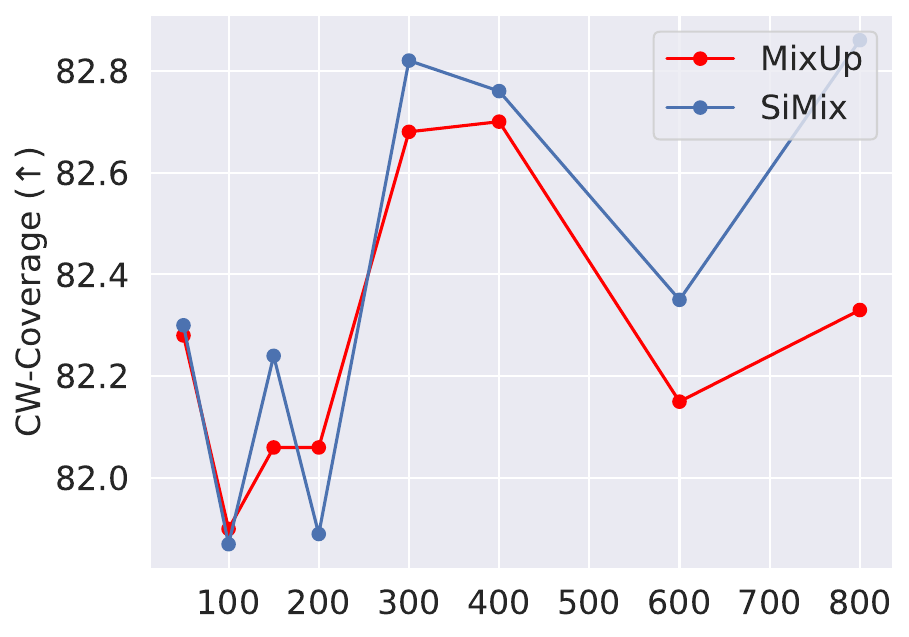}
        \label{subfig:simix_cov}
    \end{subfigure}
    \caption{Comparison of quality and diversity for two data augmentation techniques throughout discriminator training.}
    \label{fig:simix_effect}
\end{figure}
\hspace{0.9em} 
\Cref{fig:simix_effect} shows that SiMix reduces artifact in the generation process, thereby improve the sample quality.

\subsection{Class correctioness evaluation.}
\begin{figure}[h!]
    \centering
    \includegraphics[width=0.95\linewidth]{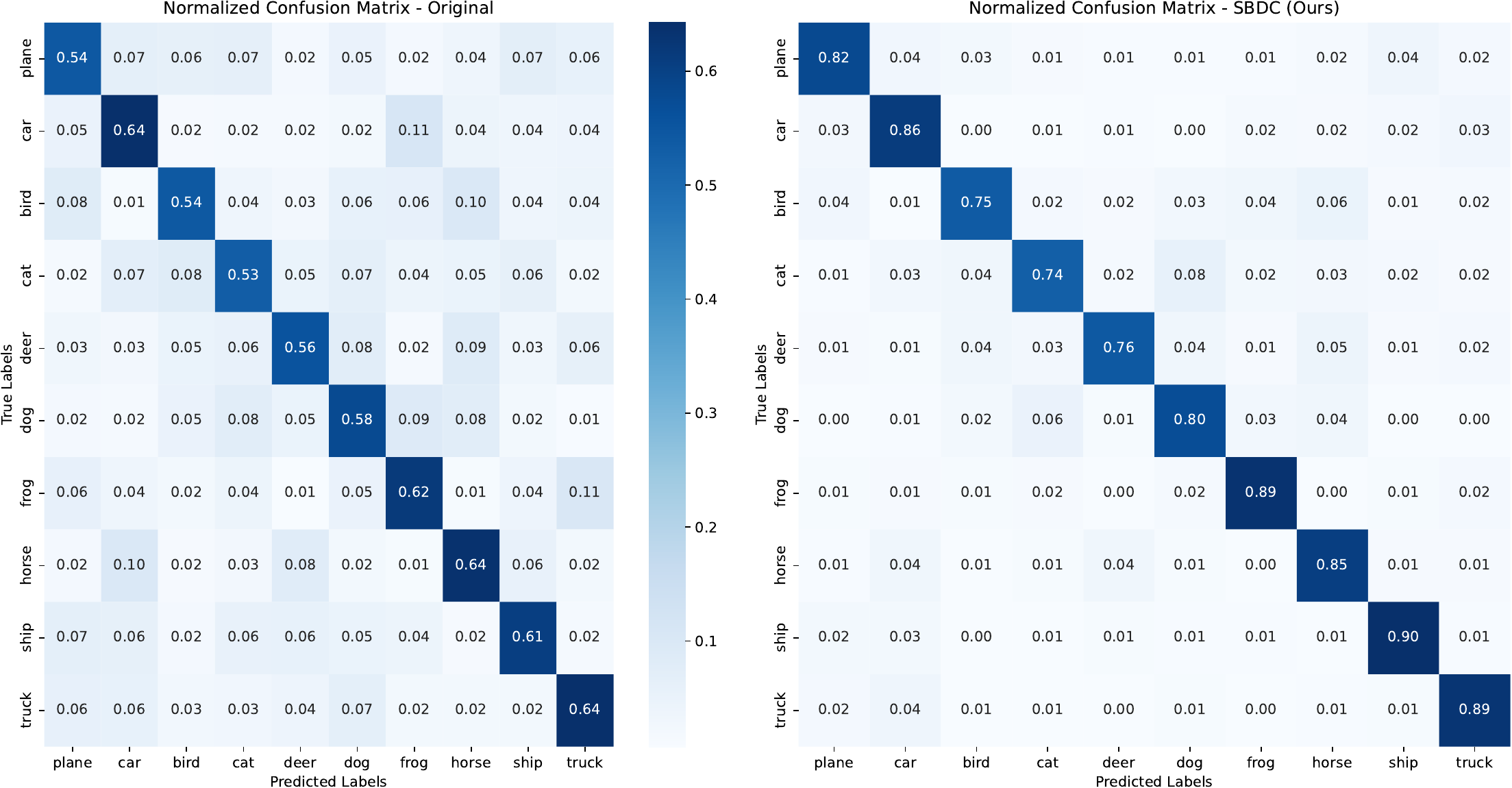}
    \caption{Confusion matrix on CIFAR10 - 40\% IDN.}
    \label{fig:confusion_matrix}
\end{figure}

\Cref{fig:confusion_matrix} shows the confusion matrix. 
SBDC indeed improves the class correctness.

\subsection{Extended Examples}
\label{subsec:visualize}

\begin{figure*}[t!]
    \centering
    \begin{subfigure}{0.8\linewidth}
        \centering
        \includegraphics[width=\linewidth]{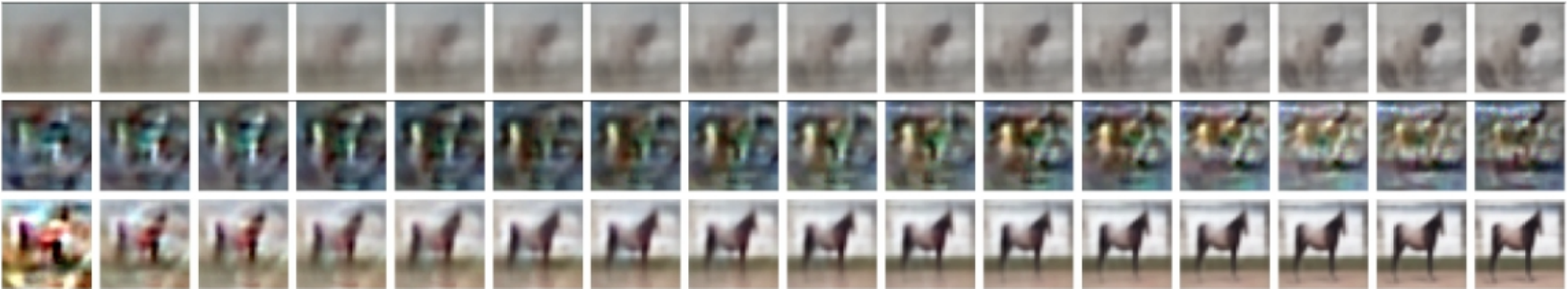}
        \caption{Instance of class "horse"}
    \end{subfigure}
    \vspace{0.4cm} 
    \begin{subfigure}{0.8\linewidth}
        \centering
        \includegraphics[width=\linewidth]{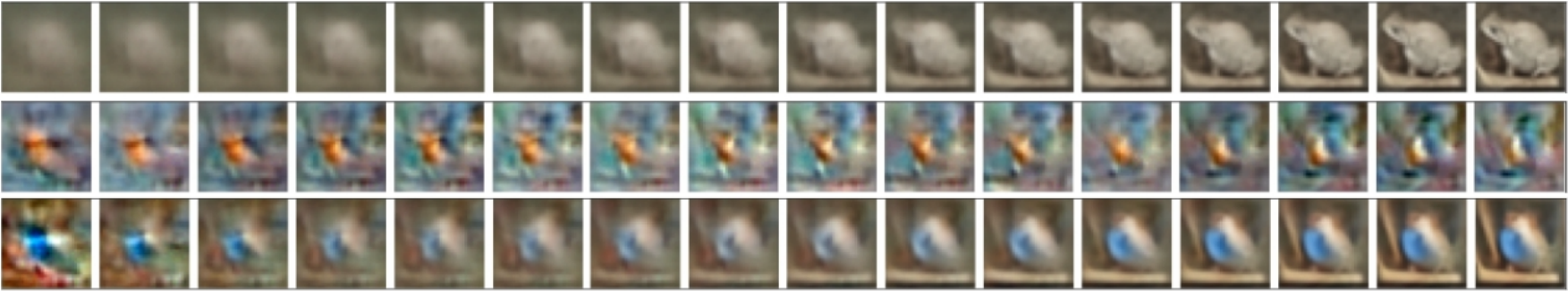}
        \caption{Instance of class "bird"}
    \end{subfigure}
    \caption{Illustration of the sampling process with correct guidance for two initial noises. 
    The first row is the origin, and the second and third rows are the discriminator gradient, and the refined denoise prediction, respectively.}
    \label{fig:illu}
\end{figure*}

\hspace{0.9em} \Cref{fig:illu} visualizes the generation process of two instances with class "horse" and "bird".

\begin{figure*}[h!]
    \centering
    \includegraphics[width=0.7\linewidth]{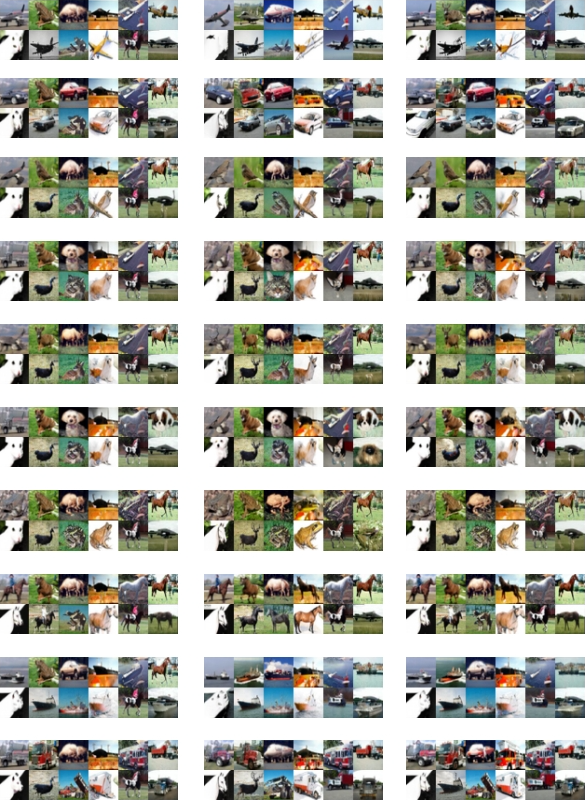}
    \caption{The uncurated generated images of baseline (left column), TDSM (middle column), and SBDC (right column) on the CIFAR-10 dataset with 50\% symmetric noise. Each block has images of the same class. The class labels are \textit{plane, car, bird, cat, deer, dog, frog, horse, ship, truck}, from top to bottom. All images are generated from the same noise.}
    \label{fig:cifar10_viz}
\end{figure*}

\begin{figure*}[h!]
    \centering
    \includegraphics[width=0.8\linewidth]{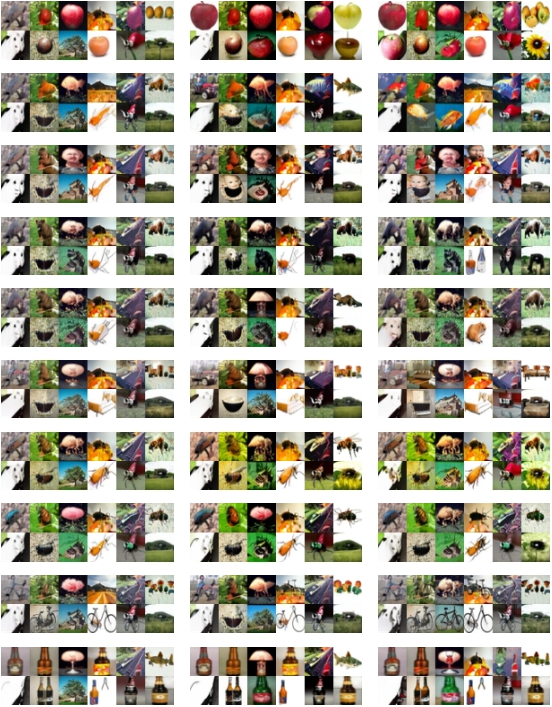}
    \caption{The uncurated generated images of baseline (left column), TDSM (middle column), and SBDC (right column) on the CIFAR-100 dataset with 40\% symmetric noise. Each block has images of the same class. The class labels are \textit{apple, aquarium fish, baby, bear, beaver, bed, bee, beetle, bicycle, bottle}, from top to bottom.}
    \label{fig:cifar100_viz}
\end{figure*}

\begin{figure*}
    \centering
    \includegraphics[width=0.8\linewidth]{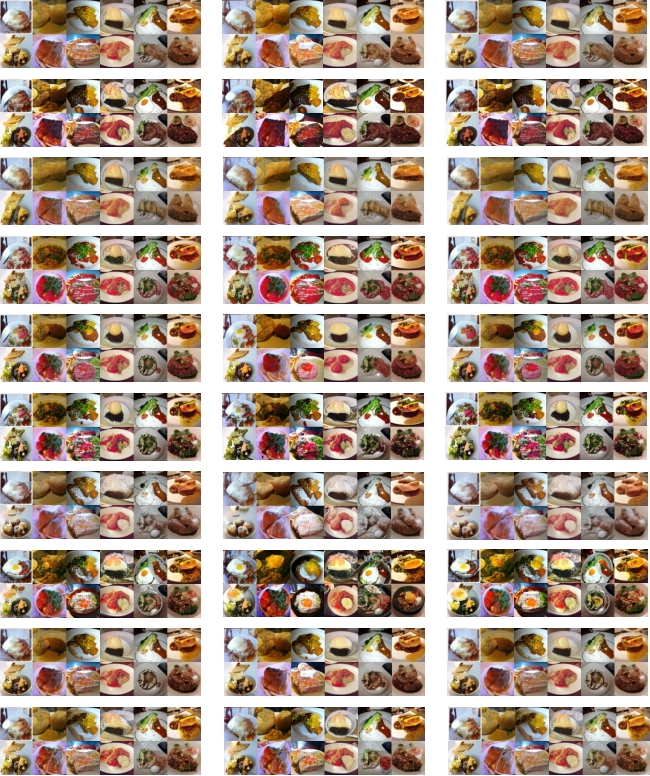}
    \caption{The uncurated generated images of baseline (left column), TDSM (middle column), and SBDC (right column) on the FOOD101 dataset. Each block has images of the same class. The class labels are \textit{apple pie, baby back ribs, baklava, beef carpaccio, beef tartare, beet salad, beignets, bibimbap, bread pudding, breakfast burrito}, from top to bottom.}
    \label{fig:food_viz}
\end{figure*}

\begin{figure*}
    \centering
    \includegraphics[width=0.8\linewidth]{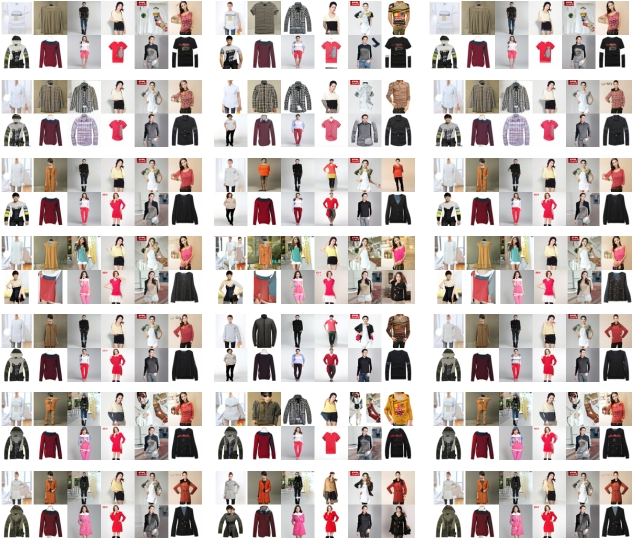}
    \caption{The uncurated generated images of baseline (left column), TDSM (middle column), and SBDC (right column) on the Clothing1M dataset. Each block has images of the same class. The class labels are \textit{t-shirt, shirt, knitwear, chiffon, sweater, hoodie, and windbreaker}, from top to bottom.}
    \label{fig:clothing_viz}
\end{figure*}

\cref{fig:cifar10_viz,fig:cifar100_viz,fig:food_viz,fig:clothing_viz} show the uncurated generated images of the baseline, TDSM, and our models. Our models exhibit proficiency in producing signals to fix generated images.

\end{appendix}

\end{document}